\documentclass[twocolumn,showpacs,superscriptaddress,nofootinbib]{revtex4}
\usepackage{graphicx} \usepackage{amsmath} \usepackage{amssymb}
\usepackage{dcolumn}
\usepackage{multirow}
\usepackage{CJK}

\newcommand{\comment}[1]{}
\newcommand{\BEQ}{\begin{equation}}
\newcommand{\EEQ}{\end{equation}}
\newcommand{\BEA}{\begin{eqnarray}}
\newcommand{\EEA}{\end{eqnarray}}
\renewcommand{\d}{{\rm d}}

\newcommand{\zmu}{\mu}

\newcommand{\w}{{\bf w}}
\newcommand{\f}{\theta}
\newcommand{\ff}{{\boldsymbol \theta}}
\newcommand{\nunu}{{\boldsymbol \nu}}

\begin{document}

\begin{CJK}{GB}{gbsn}

\title{Rank-frequency relation for Chinese characters}
\author{W.B. Deng} \affiliation{Laboratoire de Physique Statistique et
Syst\`emes Complexes, ISMANS, 44 ave. Bartholdi, Le Mans 72000,
France} \affiliation{Complexity Science Center and Institute of
Particle Physics, Hua-Zhong Normal University, Wuhan 430079,
China} \affiliation{IMMM, UMR CNRS 6283, Universit\'e du Maine,
72085 Le Mans, France}

\author{A.E. Allahverdyan} \thanks{Email:
armen.allahverdyan@gmail.com} \affiliation{Laboratoire de Physique
Statistique et Syst\`emes Complexes, ISMANS, 44 ave. Bartholdi, Le
Mans 72000, France} \affiliation{Yerevan Physics Institute,
Alikhanian Brothers Street 2, Yerevan 375036, Armenia}

\author{B. Li} \affiliation{Department of Chinese Literature, University of
Heilongjiang, Harbin 150080, China}

\author{Q. A. Wang} \affiliation{Laboratoire de Physique Statistique et
Syst\`emes Complexes, ISMANS, 44 ave. Bartholdi, Le Mans 72000,
France} \affiliation{IMMM, UMR CNRS 6283, Universit\'e du Maine,
72085 Le Mans, France}

\begin{abstract}

We show that the Zipf's law for Chinese characters perfectly holds
for sufficiently short texts (few thousand different characters).
The scenario of its validity is similar to the Zipf's law for
words in short English texts. For long Chinese texts (or for
mixtures of short Chinese texts), rank-frequency relations for
Chinese characters display a two-layer, hierarchic structure that
combines a Zipfian power-law regime for frequent characters (first
layer) with an exponential-like regime for less frequent
characters (second layer). For these two layers we provide
different (though related) theoretical descriptions that include
the range of low-frequency characters (hapax legomena). The
comparative analysis of rank-frequency relations for Chinese
characters versus English words illustrates the extent to which
the characters play for Chinese writers the same role as the words
for those writing within alphabetical systems.

\end{abstract}

\pacs{89.75.Fb, 89.75.Da, 05.65.+b}

\maketitle

\section{Introduction}

Rank-frequency relations provide a coarse-grained view on the
structure of a text: one extracts the normalized frequencies of
different words $f_1>f_2> ...$, orders them in a non-increasing
way and studies the frequency $f_r$ as a function of its rank $r$.
One widely known aspect of this rank-frequency relation that holds
for texts written in many alphabetical languages is the Zipf's
law; see \cite{wyllys,book,baa,li} for reviews,
\cite{greek,indian,lu1,Baixer} for modern instances of the law,
and \cite{zipfwiki} for extensive lists of references on the
subject. This regularity was first discovered by Estoup
\cite{est}: \BEA\label{estoup} f_r\propto r^{-\gamma} ~~ {\rm
with} ~~ \gamma\approx 1. \EEA The message of a power-law
rank-frequency relation is that there is no a single group of
dominating words in a text, they rather hold some type of
hierarchic, scale-invariant organization. This contrasts to the
exponential-like form of the rank-frequency relation that would
display a dominant group of words that is representative for the
text.

The simple form of the Zipf's law hides the mechanism behind it.
Hence there is no consensus on the origin of the law, as witnessed
by different theories proposed to explain it
\cite{mandelbrot,sole,mitra,manin,miller_li,theor,a}. An
influential group of theories explain the law from certain general
premises of the language \cite{mandelbrot,sole,mitra,manin}, e.g.
that the language trades-off between maximizing the information
transfer and minimizing the speaking-hearing effort \cite{sole},
or that the language employs its words via the optimal setting of
information theory \cite{mandelbrot}. The general problem of
derivations from this group is that explaining the Zipf's law for
the language (and verifying it for a frequency dictionary) does
not yet mean to explain the law for a concrete text, where the
frequency of the same word varies widely from one text to another
and is far from its value in a frequency dictionary.

It was held once that the Zipf's law is not especially
informative, since it is recovered by very simple stochastic
models, where words are generated through random combinations of
letters and space symbol seemingly reproducing the $f_r\propto
r^{-1}$ shape of the law \cite{miller_li}. But the reproduction is
elusive, since the model is based on features that are certainly
unrealistic for natural languages, e.g. it predicts a huge
redundancy (many words have the same frequency and length)
\cite{howes}. More recent opinions reviewed in \cite{cancho}
indicate that the Zipf's law {\it is} informative and {\it not}
reducible to any trivial statistical regularity. These opinions
are confirmed by a recent derivation of the Zipf's law from the
ideas of latent semantic analysis \cite{a}. The derivation
accounts for generalizations of the Zipf's law for high and low
frequencies, and also describes (simultaneously with the Zipf's
law) the hapax legomena effect \footnote{Hapax legomena means
literally the
  set of words that appear in the text only once. We shall employ this
  term in a broader sense as the set of words that appear few times,
  so that sufficiently many words have the same frequency. The
  description of this set is sometimes referred to as the frequency
  spectrum. }; see Appendix A for the glossary of the used linguistic
terms.

However, the Zipf's law was so far found to be absent for the
rank-frequency relation of Chinese characters
\cite{am_j,rus,dahui,sh,ha,chen_guo}, which play|sociologically,
psychologically and (to some extent) linguistically|the same role
for Chinese readers and writers as the words do in Indo-European
languages \cite{doris,chen,hoosain}.

Rank-frequency relations for Chinese characters were first studied
by Zipf and coauthors who did not find the Zipf's law \cite{zi}.
They claimed to find another power law with exponent $\gamma=2$
\cite{zi}, but this result was later on shown to be incorrect
\cite{rus}, since it was not based on any goodness of fit measure.
It was also proposed that the data obtained by Zipf are reasonably
fit with a logarithmic function $f_r=a+b\ln(c+r)$ with constant
$a$, $b$ and $c$ \cite{rus}.  The result on the absence of the
Zipf's law was then confirmed by other studies
\cite{dahui,sh,ha,chen_guo,lu}. All these authors agree that the
proper Zipf's law is absent (more generally a power law is
absent), but have different opinions on the (non-power-law) form
of the rank-frequency relation for Chinese characters: logarithmic
\cite{rus}, exponential $f_r\propto e^{-dr}$ (where $d>0$ is a
constant) \cite{dahui,sh,ha,lu} or a power-law with exponential
cutoff \cite{am_j,chen_guo}. In \cite{ckhu}, the authors describe
two different classes of rank-frequency relations for English and
Chinese literacy works, they also proposed a model to generate
such different situations.

The Zipf's law is regarded as a universal feature of human
languages on the level of words \cite{signal}
\footnote{Applications of the Zipf's law to automatic keyword
recognition are based on this fact \cite{ibm}, because keywords
are located mostly in the validity range of the Zipf's law. A
related set of applications of this law refers to distinguishing
between artificial and natural texts, fraud detection \cite{fraud}
{\it etc}; see \cite{powers} for a survey of applications in
natural language processing.}. Hence the invalidity of the Zipf's
law for Chinese characters has contributed to the ongoing debate
on controversies (coming from linguistics and experimental
psychology) on whether and to which extent the Chinese writing
system is similar to phonological writing systems
\cite{sampson,defrancis,packard}; in particular, to which extent
it is based on characters in contrast to words \footnote{We stress
already here that the Zipf's law holds for Chinese \cite{chen_guo}
and Japanese words \cite{turner}. This is expected and intuitively
follows from the possibility of literal translation from Chinese
to English, where (almost) each Chinese word is mapped to an
English one (see our glossary at Appendix A for definition of
various special terms). In this sense, the validity of the Zipf's
law for Chinese words is consistent with the validity of this law
for English texts. }.

Results reported in this work amount to the following:

-- The Zipf's law holds for sufficiently short (few thousand
different characters) Chinese texts written in Classic or Modern
Chinese \footnote{The Modern Chinese texts we studied are written
with simplified characters, while our Classic Chinese texts are
written with traditional characters. Reforms started in the
mainland China since late 1940's simplified about 2235 characters.
Traditional characters are still used officially in Hong-Kong and
Taiwan. }. Short texts are important, because they are building
blocks for understanding long texts.  For the sake of
rank-frequency relations, but also more generally, one can argue
that long texts are just mixtures (joining) of smaller,
thematically homogeneous pieces. This premise of our approach is
fully confirmed by our results.

-- The validity scenario of the Zipf's law for short Chinese texts
is basically the same as for short English texts
\footnote{\label{neo}Here and below we refer to a typical
Indo-European alphabetical based language as English, meaning that
for the sake of the present discussion differences between various
Indo-European and/or Uralic languages are not essential. Likewise,
we expect that the basic features of the rank-frequency analysis
of Chinese characters will apply for those languages (e.g.
Japanese), where the Chinese characters are used.}: the
rank-frequency relation separates into three ranges. {\it (1)} The
range of small ranks (more frequent characters) that contains
mostly function characters; we call it the pre-Zipfian range. {\it
(2)} The (Zipfian) range of middle ranks (more probable words)
that contains mostly content characters. {\it (3)} The range of
rare characters, where many characters have the same small
frequency (hapax legomena).

-- The essential difference between Chinese characters and English
words comes in for long texts, or upon mixing (joining) different
short texts. When mixing different English texts, the range of
ranks where the Zipf's law is valid quickly increases, roughly
combining the validity ranges of separate texts. Hence for a long
text the major part of the overall frequency is carried out by the
Zipfian range.  When mixing different Chinese texts, the validity
range of the Zipf's law increases very slowly. Instead there
emerges another, exponential-like regime in the rank-frequency
relation that involves a much larger range of ranks.  However, the
Zipfian range of ranks is still (more) important, since it carries
out some $40\%$ of the overall frequency.  This overall frequency
of the Zipfian range is approximately constant for all (numerous
and semantically very different) Chinese texts we studied.

-- We describe these two regimes via different (though closely
related) theories that are based on the recent approach to
rank-frequency relations \cite{a}. This description includes a
rather precise theories for rare characters (hapax legomena range)
both for long and short Chinese texts.

This work is organized as follows. The next section gives a short
introduction to Chinese characters and their differences and
similarities with English words.  Section III uncovers the Zipf's
law for short Chinese texts and compares it with the English
situation. Section IV studies the fate of the Zipf's law for long
Chinese texts. We summarize in the last section. Appendix A
contains the glossary of the used linguistic terms. Appendix B
refers to the interference experiments distinguishing between
Chinese characters and English words. Appendix C recollects
information on the studied Chinese texts. Appendix D lists the
key-characters of one studied modern Chinese text. Appendix E
reminds the Kolmogorov-Smirnov test that is employed for checking
the quality of our numerical fitting.

\section{Chinese characters versus English words} \label{intro}

Here we shortly remind the main differences and similarities
between Chinese characters and English words; see Footnote
\ref{neo} in this context. This subject generated several
controversies (myths as it was put in \cite{defrancis}), even
among expert sinologists
\cite{chen,hoosain,hoosain_1,sampson,defrancis,packard}.

This section is not needed for presenting our results (hence it
can be skipped upon first reading), but is necessary for a deeper
understanding of our results and motivations.

The main qualitative conclusion of this section is that in
contrast to English words, Chinese characters have generally more
different meanings, they are more flexible, they could combine
with other characters to convey different specific meanings. So
there are characters, which appear many times in the text, but
their concrete meanings are different in different places.

{\bf 1.} The unit of Chinese writing system is the character: a
spatially marked pattern of strokes phonologically realized as a
single syllable (please consult Appendix A for a glossary of
various linguistic terms used in the paper). Generally, each
character denotes a morpheme or several different morphemes.

{\bf 2.} The Chinese writing system evolved by emphasizing the
concept of the character-morpheme, to some extent blurring the
concept of the multi-syllable word. In particular, spaces in the
Chinese writing system are put in between of characters and not in
between of multi-syllable words \footnote{An immediate question is
whether Chinese readers will benefit from reading a
character-written text, where the words boundaries are indicated
explicitly. For normal sentences the readers will not benefit,
i.e. it does not matter whether the word boundaries are indicated
explicitly or not \cite{liu}. But for difficult sentences the
benefit is there \cite{huang}.}. Thus a given sentence can have
different meanings when being separated into different sequences
of words \cite{hoosain_1}, and parsing a string of Chinese
characters into words became a non-trivial computational problem;
see \cite{luo} for a recent review.

{\bf 3.} Psycholinguistic research shows that the characters are
important cognitive and perceptual units for Chinese writers and
readers \cite{chen,doris,hoosain}, e.g. Chinese characters are
more directly related to their meanings than English words to
their meanings \cite{hoosain} \footnote{To get a fuller picture of
this effect let us denote $\tau_f(E)$ and $\tau_f(C)$ for English
and Chinese phonology activation times, respectively, while
$\tau_m(E)$ and $\tau_m(C)$ stand for respective meaning
activation times. The phonology activation time is the time passed
between seeing a word in English (or character in Chinese) and
pronouncing it; likewise, for the meaning activation time. Now
these quantities hold \cite{hoosain}: $\tau_f(E)< \tau_m(C)\simeq
\tau_f(C)< \tau_m(E)$.}; see Appendix B for additional details.
The explanation of this effect would be that characters (compared
to English words) are perceived holistically as a meaning-carrying
objects, while English words are yet to be reconstructed from a
sequence of their constituents (phonemes and syllables)
\footnote{A simpler explanation would be that the characters are
perceived as pictograms directly pointing to their meaning. In its
literal form this explanation is not correct, since
characters-pictograms are not frequent in Chinese
\cite{defrancis,sceptic}. }.

\comment{ {\bf 3.1} The word inferiority effect (see its
description in Appendix B2) demonstrates that the perception of
Chinese characters is not similar to that of English letters
\cite{chen}, and the perception of Chinese words is not similar to
the perception of English words \cite{chen}. }

{\bf 4.} One-character words dominate in the following specific
sense. Some 54\% of modern Chinese word {\it tokens} are
single-character, two-character word tokens amount to 42\%; the
remaining words have three or more characters \cite{chen_1993}.
For modern Chinese word {\it types} the situation is different:
single character words amount to some 10\% against 66\% of
two-character words \cite{chen_1993}. Classic Chinese texts have
more single-character words (tokens), the percentage varies
between some 60\% and 80\% for texts written in different periods.

The modern Chinese has $\approx 10440$ basic (root) morphemes. 93
\% of them are represented by single characters. The overall
number of Chinese characters is $\approx 18 000$.

{\bf 5.} A minor part of multi-character words are multi-character
morphemes, i.e. their separate characters do not normally appear
alone (they are fully bound).  Examples of this are the
two-character Chinese words for {\it grape} ``ÆÏÌÑ" {\it (p\'u
t\'ao)}, {\it dragonfly} ``òßòÑ" {\it (q\=\i ng \ t\'\i ng)}, {\it
olive} ``éÏé­" {\it (g\v an l\v an)}. Estimates show that some
10\% of all characters are fully bound \cite{defrancis}.

A related set of examples is provided by two-character words,
where the separate characters do have an independent meaning, but
this meaning is not directly related to the meaning of the word,
e.g. ``¶«Î÷" {\it (d\=ong x\=\i)} means {\it thing}, but literally
it amounts to {\it east-west}, or ``ÊÖ×ã" {\it (sh\v ou z\'u)}
means {\it close partnership}, but literally {\it hand-foot}.)

{\bf 6.} The majority of the multi-character words are semantic
compounds: their separate characters can stand alone and are
related to the overall meaning of the word. Importantly, in most
cases, the separate meanings of the component characters are wider
than the (relatively unique) meaning of the compound two-character
word. An example of this situation is the two-character Chinese
word for {\it train} ``»ð³µ" {\it (hu\v o \ ch\=e)}: its first
character ``»ð" {\it (hu\v o)} has the meaning of {\it fire, heat,
popular, anger, etc}, while the second character ``³µ" {\it
(ch\=e)} has the meaning of {\it vehicle, machine, wheeled, lathe,
castle, etc}.

Note that in Chinese there is a certain freedom in grouping
morpheme into different combinations. Hence it is not easy to
distinguish the semantic compounds from lexical phrases.


{\bf 7.} At this point we shall argue that in general Chinese
characters have a larger number of different meanings than English
words. This statement will certainly appear controversial, if it
is taken without proper caution, and is explained without proper
usage of linguistic terms (see our glossary at Appendix A);
consult Footnote \ref{garush} in this context.

First of all note the difference between polysemes and homographs:
polysemes are two related meanings of the same character (word),
homographs are two characters (words) that are written in the same
way, but their meanings are far from each other \footnote{Note
that polysemes are defined to be related meanings of the same
word, while homographs are defined to be different words. This is
natural, but also to some extent conventional, e.g. one can still
define homographs as far away meanings of the same word.}. Now
many characters are simultaneously homographs and polysemes, e.g.
character ``Ã÷" {\it (m\'\i ng)} means {\it brilliant, light,
clear, next, etc}. Here the first three meanings are related and
can be viewed as polysemes.  The fourth meaning {\it next} is
clearly different from the previous three.  Hence this is a
homograph. Another example is the character ``·¢" {\it (f\=a or
f\`a)} that can mean {\it hair, send out, fermentation, etc}. All
these three meanings are clearly different; hence we have
homographs. Note the following peculiarity of the above two
examples: the first example is a non-heteronym (homophonic)
character, i.e.  it is read in the same way irrespectively whether
it means {\it light} or {\it next}.  The second example is a
heteronym character: it written in the same way, but is read
differently depending on its meaning.

In most cases, heteronym characters|those which are written in the
same way, but have different pronunciations|have at least two
sufficiently different meanings.  The disambiguation of their
meaning is to be provided by the context of the sentence and/or
the shared experience of the writer and reader \footnote{Note that
homophony in Chinese is much larger than homography: in average a
syllable has around 12--13 meanings \cite{doris}. Hence, in a
sense, characters help to resolve the homophony of Chinese speech.
This argument is frequently presented as an advantage of the
character-based writing system, though it is not clear whether
this system is here not solving the problem that was invited by
its usage \cite{sceptic}. }.

Surely, also English words can be ambiguous in meaning (e.g. {\it
get} means {\it obtain}, but also {\it understand} = {\it have
knowledge}), but there is an essential difference.  The major
contribution of the meaning ambiguity in English is the polysemy:
one word has somewhat different, but also closely related
meanings. In contrast, many Chinese characters have widely
different meanings, i.e. they are homographs rather than
polysemes.

However, we are not aware of any quantitative comparison between
homography of Chinese versus English. This may be related to the
fact that it is sometimes not easy to distinguish between polysemy
and homophony (see the glossary in Appendix A). Still the above
statement on Chinese characters having a larger number of
different meanings can be quantitatively illustrated via the
relative prevalence of heteronyms in Chinese. The amount of
heteronyms in English is negligible, e.g. in rather complete list
of heteronyms presented in \cite{english_homographs}, we noted
only 74 heteronyms \footnote{Not counting those heteronyms that
arise because an English word happens to coincide with a foreign
special name, e.g. {\it Nancy} [English name] and {\it Nancy} city
in France.}, and only three of them had more than 2 meanings. This
is a tiny amount of the overall number of English words ($>
5\times 10^5$).  To compare this with the Chinese situation, we
note that at least some 14\% of modern Chinese and 25\% of
traditional characters are heteronyms, which normally have at
least two widely different meanings. Within the most frequent 5700
modern characters the number of heteronyms is even larger and
amounts to 22 \% \cite{chen_1993} \footnote{\label{garush}One
should not conclude that in average the Chinese character has more
meanings than the English word, because there is a large number of
characters|between 10 and 14 \% depending on the type of the
dictionary employed \cite{obukhova}|that do not have lexical
meaning, i.e. they are either function words (grammatical meaning
mainly) or characters that cannot appear alone (bound characters).
If now the number of meanings for each character is estimated via
the number of entries in the explanatory dictionary|which is more
or less traditional way of searching for the number of meanings,
though it mixes up homography and polysemy|the average number of
meanings per a Chinese character appears to be around 1.8--2
\cite{obukhova}. This is smaller than the average number of
(necessarily polysemic) meanings for an English word that amounts
to 2.3.}.

{\bf 8.} Chinese nouns are generally less abstract: whenever
English creates a new word via conceptualizing the existing one,
Chinese tends to explain the meaning via using certain basic
characters (morphemes). Several basic examples of this scenario
include: length=long+short ``³¤¶Ì" {\it (ch\'ang du\v an)},
landscape=mountains+water ``ɽˮ" {\it (sh\=an shu\v \i)},
adult=big+person ``´óÈË" {\it (d\`a r\'en)},
population=person+mouth ``ÈË¿Ú" {\it (r\'en k\v ou)},
astronomy=heaven+script ``ÌìÎÄ" {\it (ti\=an w\'en)},
universe=great+emptiness ``Ì«¿Õ" {\it (t\`ai k\=ong)}. English
tools for making abstract words include prefixes, {\it poly-},
{\it super-}, {\it pro-}, {\it etc} and suffixes, {\it -tion},
{\it -ment}. These tools either do not have Chinese analogs, or
their usage can generally be suppressed.

English words have inflections to indicate the tense of verbs, the
number for nouns or the degree for adjectives. Chinese characters
generally do not have such linguistic attributes \footnote{Chinese
expresses temporal ordering via context, e.g. adding words {\it
tomorrow} or {\it yesterday}, or by aspects. The difference
between tense and aspect is that the former implicitly assumes an
external observer, whose reference time is compared with the time
of the event described by the sentence. Aspects order events
according to whether they are completed, or to which extent they
are habitual. Indo-European languages tie up tense and aspect. The
tie is weaker for Slavic Indo-European languages. Chinese has
several tenses including perfective, imperfective and neutral.  },
their role is carried out by the context of the sentence(s)
\footnote{Chinese has certain affixes, but they can be and are
suppressed whenever the issue is clear from the context. }.

To summarize this section, the differences between Chinese and
English writing systems can be viewed in the context of the two
features: emphasizing the role of base (root) morphemes and
delegating the meaning to the context of the sentence whenever
this is possible \cite{doris}.

The quantitative conclusion to be drawn from the above discussion
is that Chinese characters have more different meanings, they are
flexible, they could combine with other characters to convey
different specific meanings. Anticipating our results in the
sequel, we expect to see a group of characters, which appear many
times in the text, but their concrete meanings are different in
different places of the text.

\begin{table*}
\begin{center}
\tabcolsep0.06in \arrayrulewidth0.5pt
\renewcommand{\arraystretch}{1.2}
\caption{\label{tab_1} Parameters of the modern Chinese texts (see
Appendix C for further details). $N$ is the total number of
characters in the text. The number of different characters is $n$.
The Zipf's law $f_r=cr^{-\gamma}$ holds for the ranks $r_{\rm
min}\leq r\leq r_{\rm max}$; see section III A. Here $\sum_{k <
r_{\rm min}}f_k$ and $\sum_{k =r_{\rm min}}^{r_{\rm max}}f_k$ are
the total frequencies carried out by the pre-Zipfian and Zipfian
domain, respectively.
\newline $d$ is the difference between the total frequency of the
Zipfian domain got empirically and its value according to the
Zipf's law: $d=\sum_{k=r_{\rm min}}^{r_{\rm max}}
(ck^{-\gamma}-f_k)$. Its absolute value $d$ characterizes the
global precision of the Zipf's law. \newline AQZ \& KLS means
joining the texts AQZ and KLS. \newline $r_b$ is the conventional
borderline rank between the exponential-like range and the hapax
legomena; see section IV B 2 for its definition. Whenever we put
``-" instead of it, we mean that either the exponential-like range
is absent or it is not distinguishable from the hapax legomena.}
\vskip 0.3cm
\begin{tabular}{|c|c|c|c|c|c|c|c|c|c|c|c|c|}
\hline
Texts & $N$ &  $n$ &$r_{\rm min} $& $r_{\rm max}$ & $c$ & $\gamma$ & $\sum_{k < r_{\rm min}}f_k$ & $\sum_{k =r_{\rm min}}^{r_{\rm max}}f_k$  & $\vert$$d$$\vert$ & $r_b$   \\
 \hline
AQZ & 18153 & 1553 & 56 & 395 & 0.2239 & 1.03 & 0.42926 & 0.38424  & 0.00624 & -\\
KLS & 20226 & 2047 & 62 & 411 & 0.169 & 0.97 & 0.39971 & 0.379728  & 0.005728 &-\\
AQZ \& KLS & 38379 & 2408 & 66 & 439 & 0.195 & 1.0 & 0.41684 & 0.369  & 0.0022 &-\\
PFSJ & 705130 & 3820 & 67 & 583 & 0.234 & 1.03 & 0.39544 & 0.425379  & 0.00842 &1437\\
 SHZ & 704936 & 4376 & 78 & 590 & 0.225 & 1.02 & 0.39905 & 0.42  & 0.009561 &1618\\
\hline
\end{tabular}
\end{center}
\end{table*}

\begin{table*}[ht]
\begin{center}
\tabcolsep0.06in \arrayrulewidth0.5pt
\renewcommand{\arraystretch}{1.2}
\caption{\label{tab_2} Parameters of classic Chinese texts (see
Appendix C for further details). Notations have the same meaning
as in Table \ref{tab_1}. Here 4 texts means joining of the texts
CQF, SBZ, WJZ, and HLJ. Also, 7 (10,14) texts mean joining of the
4 with other 3 (6,11) classic texts, which we do not mention
separately, because they give no new information. }  \vskip 0.3cm
\begin{tabular}{|c|c|c|c|c|c|c|c|c|c|c|c|c|}
 \hline
Texts & $N$ &  $n$ &$r_{\rm min} $& $r_{\rm max}$ & $c$ & $\gamma$ & $\sum_{k < r_{\rm min}}f_k$ & $\sum_{k =r_{\rm min}}^{r_{max}}f_k$  & $\vert$$d$$\vert$ & $r_b$ \\
\hline
CQF & 30017 & 1661 & 47 & 365 & 0.1778 & 0.985 & 0.43906 & 0.38997  & 0.00441 &-\\
SBZ & 24634 & 1959 & 52 & 357 & 0.1819 & 0.972 & 0.42828 & 0.408787  & 0.004353 &-\\
WJZ & 26330 & 1708 & 46 & 360 & 0.208 & 0.999 & 0.40434 & 0.418733  & 0.006923 &-\\
HLJ & 26559 & 1837 & 56 & 372 & 0.209 & 1.01 & 0.43674 & 0.379454  & 0.000832 &-\\
CQF \& SBZ & 54651 & 2528 & 68 & 483 & 0.19498 & 0.989 & 0.42031 & 0.401661  & 0.00483 &-\\
CQF \& WJZ & 56347 & 2302 & 66 & 439 & 0.20654 & 1.002 & 0.42815 & 0.383514  & 0.00564 &-\\
CQF \& HLJ & 56576 & 2458 & 65 & 416 & 0.19498 & 0.998 & 0.43138 & 0.38654  & 0.00913 &-\\
SBZ \& WJZ & 50964 & 2505 & 68 & 465 & 0.20512 & 0.992 & 0.40116 & 0.409017 & 0.00382 &-\\
SBZ \& HLJ & 51193 & 2608 & 72 & 423 & 0.20893 & 1.000 & 0.41157 & 0.369598  & 0.00798 &-\\
WJZ \& HLJ & 52889 & 2303 & 66 & 432 & 0.23988 & 1.035 & 0.43044 & 0.380801  & 0.002321 &-\\
4 texts & 107540 & 3186 & 75 & 528 & 0.22387 & 1.021 & 0.42526 & 0.391818  & 0.0007 & 681\\
7 texts & 190803 & 4069 & 57 & 513 & 0.158 & 0.97 & 0.39381 & 0.4102 & 0.00331 & 789\\
 10 texts & 278557 & 4727 & 67 & 552 & 0.168 & 0.978 & 0.38058 & 0.4015 & 0.00217 & 1015\\
14 texts & 348793 & 5018 & 78 & 625 & 0.176 & 0.98 & 0.39116 & 0.418983  & 0.00954 & 1223\\
 SJ & 572864 & 4932 & 76 & 535 & 0.236 & 1.025 & 0.40153 & 0.41253  & 0.007564 & 1336\\
\hline
\end{tabular}
\end{center}
\end{table*}

\begin{table*}
\begin{center}
\tabcolsep0.08in \arrayrulewidth0.5pt
\renewcommand{\arraystretch}{1.2}
\caption{ \label{tab_3} Parameters of four English texts and their
mixtures: {\it The Age of Reason} (AR) by T. Paine, 1794 (the
major source of British deism). {\it Time Machine} (TM) by H. G.
Wells, 1895 (a science fiction classics). {\it Thoughts on the
Funding System and its Effects} (TF) by P. Ravenstone, 1824
(economics). {\it Dream Lover} (DL) by J. MacIntyre, 1987 (a
romance novella). TF \& TM means joining the texts TF and TM.
\newline The total number of words $N$, the number of different
words $n$, the lower $r_{\rm min}$ and the upper $r_{\rm max}$
ranks of the Zipfian domain, the fitted values of $c$ and
$\gamma$, the overall frequencies of the pre-Zipfian and Zipfian
range, and the difference $d$ between the total frequency of the
Zipfian domain got empirically and its value according to the
Zipf's law: $d=\sum_{k=r_{\rm min}}^{r_{\rm max}}
(ck^{-\gamma}-f_k)$.} \vskip 0.3cm
\begin{tabular}{|c|c|c|c|c|c|c|c|c|c|c|}
\hline Texts & $N$ &  $n$ &$r_{\rm min} $& $r_{\rm max}$ & $c$ &
$\gamma$ & $\sum_{k < r_{\rm min}}f_k$& $\sum_{k =r_{\rm
min}}^{r_{\rm max}}f_k$ &
$\vert$$d$$\vert$   \\
\hline
TF & 26624 &  2067 &  36   & 371 & 0.168& 1.032 & 0.44439   & 0.35158   & 0.00333\\
TM & 31567 & 2612 & 42&332 &0.166 & 1.041 & 0.45311   &  0.33876  & 0.01004\\
AR & 22641 & 1706 &  32   & 339& 0.178& 1.038 &  0.47254  &  0.33947  & 0.00048\\
DL& 24990 & 1748 &  34  & 230& 0.192 & 1.039 &  0.47955  &  0.33251  & 0.02145\\
TF \& TM & 54191 &  3408 &  30& 602 & 0.139 & 1.013 &  0.43508  &  0.40876  & 0.02091\\
TF \& AR & 45265 & 2656 &  33 & 628 & 0.138 & 0.998 &  0.45468  &  0.41045  & 0.00239\\
TF \& DL & 47614 & 2877 &  28 & 527& 0.162& 1.014 &  0.42599  &   0.42261 & 0.01490\\
TM \& AR & 54208 & 3184 &  43 & 592 &0.157 & 1.021 &  0.47582  &  0.39687  & 0.00491\\
TM \& DL & 56557 & 3154 &  45 & 493 & 0.161& 1.023 &  0.46726  &  0.38456  & 0.01211\\
AR \& DL & 47631 & 2550 &  38 & 496  & 0.165& 1.012 & 0.45375   &  0.39236  & 0.00947\\
Four texts & 101822& 4047& 39 & 927 & 0.158 &  1.015  &  0.44245  & 0.44158   & 0.00187\\
\hline
\end{tabular}
\end{center}
\end{table*}

\section{The Zipf's law for short texts} \label{short}

We studied several Chinese and English texts of different lengths
and genres written in different epochs; see Tables \ref{tab_1},
\ref{tab_2} and \ref{tab_3}. Some Chinese texts were written using
modern characters, others employ traditional Chinese characters;
see Tables \ref{tab_1} and \ref{tab_2}. Chinese texts are
described in Appendix C. English texts are described in Table
\ref{tab_3}. The texts can be classified as short (total number of
characters or words is $N=1-3\times 10^4$) and long ($N> 10^5$).
They generally have different rank-frequency characteristics, so
discuss them separately.

For fitting empiric results we employed the linear least-square
method (linear fitting), but the also checked its results with
other methods (KS test, non-linear fitting and the maximum
likelihood method).  We start with a brief remainder of the linear
fitting method.

\subsection{Linear fitting} \label{linear}

For each Chinese text we extract the ordered frequencies of
different characters [the number of different characters is $n$;
the overall number of characters in a text is $N$]: \BEA
\label{00} \{f_r\}_{r=1}^{n}, ~~ f_1\geq ...\geq f_{n}, ~~
{\sum}_{r=1}^{n} f_r =1. \EEA Exactly the same method is applied
to English texts for studying the rank-frequency relation of
words.

\begin{figure}
\begin{center}
\resizebox{0.8\columnwidth}{!}{%
 \includegraphics{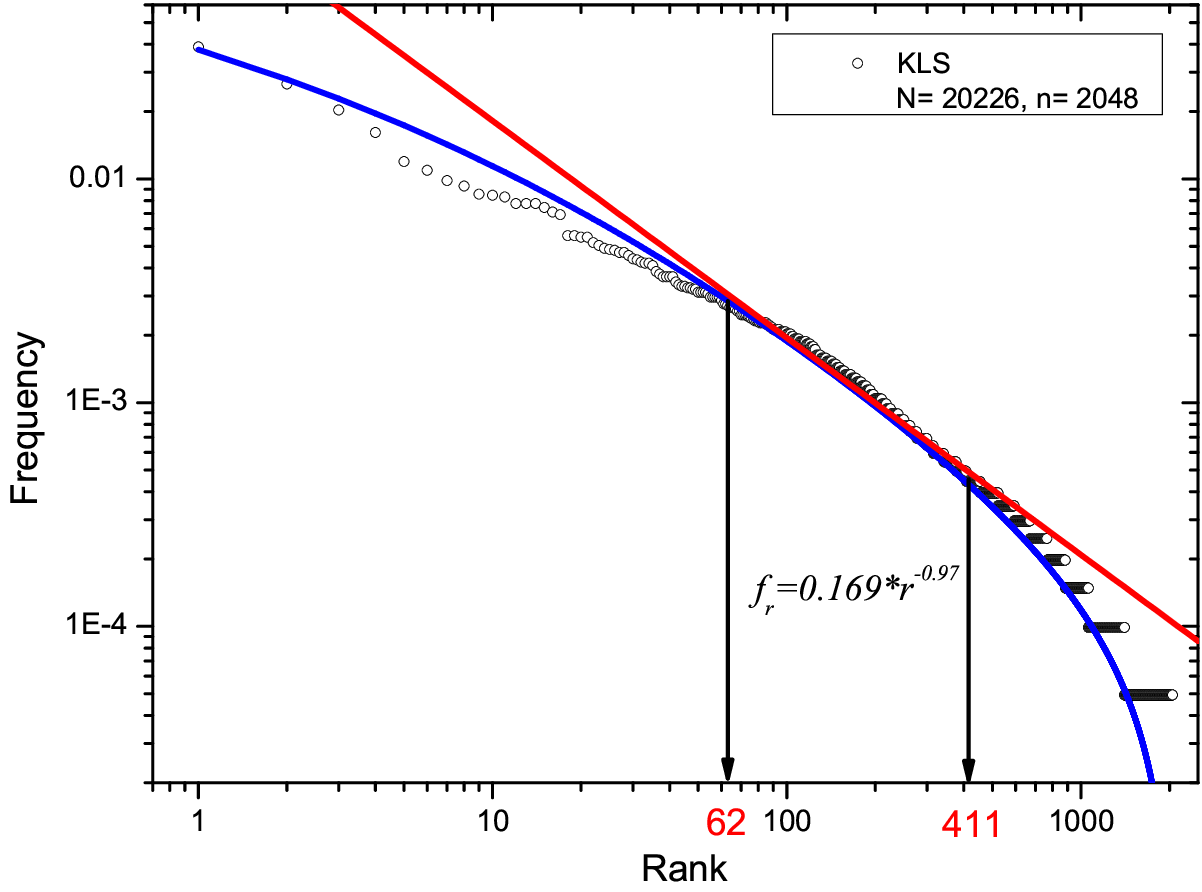} }
\caption{\label{fig_1} (Color online) Frequency versus rank for
the short modern Chinese text KLS; see Appendix C for its
description. Red line: the Zipf curve $f_r=0.169 r^{-0.97}$; see
Table \ref{tab_1}. Arrows and red numbers indicate on the validity
range of the Zipf's law. Blue line: the numerical solution of
(\ref{g4}, \ref{g5}) for $c=0.169$. It coincides with the
generalized Zipf law (\ref{g6}) for $r>r_{\rm min}=62$. The
step-wise behavior of $f_r$ for $r>r_{\rm max}$ refers to hapax
legomena. }
\end{center}
\end{figure}

We fit the data $\{f_r\}_{r=1}^{n}$ with a power law:
$\hat{f}_r=cr^{-\gamma}$. Hence we represent the data as \BEA
\{y_r(x_r)\}_{r=1}^{n}, ~~ y_r=\ln f_r, ~~ x_r=\ln r,
\label{kabanyan} \EEA and fit it to the linear form
$\{\hat{y}_r=\ln c-\gamma x_r\}_{r=1}^{n}$. Two unknowns $\ln c$
and $\gamma$ are obtained from minimizing the sum of squared
errors [linear fitting] \BEA SS_{\rm err}={\sum}_{r=1}^{n}
(y_r-\hat{y}_r)^2. \EEA It is known since Gauss that this
minimization produces \BEA \label{laude} -\gamma^* = \frac{
\sum_{k=1}^n (x_k-\overline{x})(y_k-\overline{y})    }{
\sum_{k=1}^n (x_k-\overline{x})^2}, ~~ \ln
c^*=\overline{y}+\gamma^*\overline{x}, \EEA where we defined \BEA
\overline{y}\equiv \frac{1}{n}{\sum}_{k=1}^n y_k, ~~
\overline{x}\equiv \frac{1}{n}{\sum}_{k=1}^n x_k. \EEA As a
measure of fitting quality one can take: \BEA {\rm
min}_{c,\gamma}[SS_{\rm err}(c,\gamma)]=SS_{\rm
err}(c^*,\gamma^*)=SS_{\rm err}^*. \EEA This is however not the
only relevant quality measure. Another (more global) aspect of
this quality is the coefficient of correlation between
$\{y_r\}_{r=1}^{n}$ and $\{\hat{y}_r\}_{r=1}^{n}$ \cite{book,lili}
\BEA R^2=\frac{  \left[\, \sum_{k=1}^n (y_k-\bar{y})
(\hat{y}^*_k-\overline{\hat{y}^*})\, \right]^2 }{\sum_{k=1}^n
(y_k-\bar{y})^2 \sum_{k=1}^n
(\hat{y}^*_k-\overline{\hat{y}^*})^2}, \EEA where \BEA
\hat{y}^*=\{\hat{y}^*_r=\ln c^*-\gamma^* x_r\}_{r=1}^{n}, ~~~
\overline{\hat{y}^*}\equiv \frac{1}{n}{\sum}_{k=1}^n \hat{y}^*_k.
\EEA For the linear fitting (\ref{laude}) the squared correlation
coefficient is equal to the coefficient of determination, \BEA
R^2= {\sum}_{k=1}^n (\hat{y}^*_k-\overline{y})^2\left/
{\sum}_{k=1}^n(y_k-\overline{y})^2, \right. \EEA the amount of
variation in the data explained by the fitting \cite{book,lili}.
Hence $SS_{\rm err}^*\to 0$ and $R^2\to 1$ mean good fitting.  We
minimize $SS_{\rm err}$ over $c$ and $\gamma$ for $r_{\rm min}\leq
r\leq r_{\rm max}$ and find the maximal value of $r_{\rm
max}-r_{\rm min}$ for which $SS_{\rm err}^*$ and $1-R^2$ are
smaller than, respectively, $0.05$ and $0.005$. This value of
$r_{\rm max}-r_{\rm min}$ also determines the final fitted values
$c^*$ and $\gamma^*$ of $c$ and $\gamma$, respectively; see Tables
\ref{tab_1}, \ref{tab_2}, \ref{tab_3} and Fig.~\ref{fig_1}. Thus
$c^*$ and $\gamma^*$ are found simultaneously with the validity
range $[r_{\rm max},r_{\rm max}]$ of the law. Whenever there is no
risk of confusion, we for simplicity refer to $c^*$ and $\gamma^*$
as $c$ and $\gamma$, respectively.

\subsection{Empiric results on the Zipf's law}

Here are results produced via the above linear fitting.

{\bf 1.} For each Chinese text there is a specific (Zipfian) range
of ranks $r\in [r_{\rm min}, r_{\rm max}]$, where the Zipf's law
$f_r=cr^{-\gamma}$ holds with $\gamma\approx 1$ and $c\lesssim
0.25$; see Tables \ref{tab_1}, \ref{tab_2} and Fig.~\ref{fig_1}.
Both for $r<r_{\rm min}$ and $r>r_{\rm max}$ the frequencies are
below the Zipf curve; see Fig.~\ref{fig_1}. A power rank-frequency
relation with exponent $\gamma\approx 1$ is the hallmark of the
Zipf's law \cite{wyllys,book,baa,li}.

\begin{figure}
\begin{center}
\resizebox{0.8\columnwidth}{!}{%
 \includegraphics{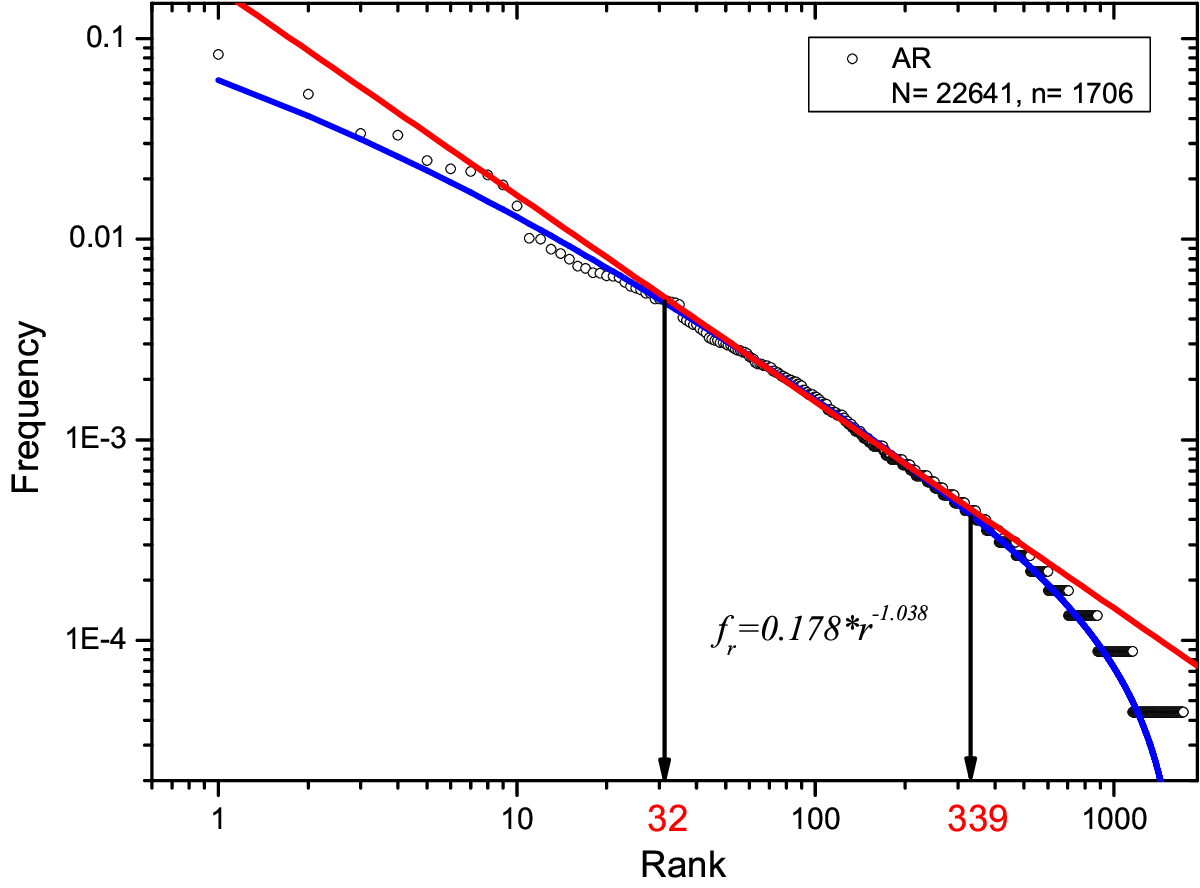} }
\caption{ \label{fig_2} (Color online) Frequency vs. rank for the
English text AR; see Table \ref{tab_3}. Red line: the Zipf curve
$f_r=0.178 r^{-1.038}$. Other notations have the same meanings as
in Fig.~1. }
\end{center}
\end{figure}

Note that though the validity range $|r_{\rm max}- r_{\rm min}|$
is few times smaller than the maximal rank $n$ (see Tables
\ref{tab_1} and \ref{tab_2} and Figs.~\ref{fig_1} and
\ref{fig_2}), it is relevant, since it contains a sizable amount
of the overall frequency: for Chinese texts (short or long) the
Zipfian range carries 40 \% of the overall frequency, i.e.
$\sum_{k=r_{\rm min}}^{r_{\rm max}} f_k\simeq 0.4$.

{\bf 2.} In the pre-Zipfian range $1\leq r<r_{\rm min}$ the
overall number of function and empty characters is more than the
number of content characters.  Function and empty characters serve
for establishing grammatical constructions (e.g. ``µÄ" {\it (de)},
``ÊÇ" {\it (sh\`\i)}, ``ÁË" {\it (le)}, ``²»" {\it (b\`u)}, ``ÔÚ"
{\it (z\`ai)}). (We shall list them separately, though for our
purposes they can be joined together; the main difference between
them is that the empty characters are not used alone.)

But the majority of characters in the Zipfian range do have a
specific meaning (content characters). A subset of those content
characters has a meaning that is specific for the text and can
serve as its key-characters; see Appendix D and Table \ref{tab_7}
for an example.

Let us take for an example the modern Chinese text KLS; see Table
\ref{tab_1} (this text concerns military activities; see Appendix
C). The pre-Zipfian range of this text contains 61 characters.
Among them there are, 24 function characters, 9 empty characters,
25 content characters, and finally there are 3 key-characters
\footnote{We present that meaning of the character which is most
relevant in the context of the text.}: horn ``ºÅ" {\it (h\`ao)},
army ``¾ü" {\it (j\=un)} and soldier ``±ø" {\it (b\=\i n)}.

The Zipfian range of the KLS contains 350 characters. Among them,
91 are function, 10 are empty, 230 are content and 19 are
key-characters (see Appendix D for the full list of key-characters
for this text).

{\bf 3.} The absolute majority of different characters with ranks
in $[r_{\rm min}, r_{\rm max}]$ have different frequencies.  Only
for $r\simeq r_{\rm max}$ the number of different characters
having the same frequency is $\simeq 10$. For $r> r_{\rm max}$ we
meet the hapax legomena effect: characters occurring only few
times in the text (i.e. $f_rN=1,2,3...$ is a small integer), and
many characters having the same frequency $f_r$ \cite{baa}. The
effect is not described by any smooth rank-frequency relation,
including the Zipf's law. Hence for short texts we get that the
Zipf's law holds for as high ranks as possible, in the sense that
for $r> r_{\rm max}$ no smooth rank-frequency relations are
possible at all.

Note that the very existence of hapax legomena is a non-trivial
effect, since one can easily imagine (artificial) texts, where
(say) no character appear only once.  The theory reviewed below
allows to explain the hapax legomena range together with the
Zipf's law; see below. It also predicts a generalization of the
Zipf's law to frequencies $r< r_{\rm min}$ that is more adequate
(than the Zipf's law) to the empiric data; see Figs.~\ref{fig_1}
and \ref{fig_2}.

{\bf 4.} All the above results hold for relatively short English
\cite{a}; see Table \ref{tab_3} and Fig.~\ref{fig_2}. In
particular, the Zipfian range of English texts also contains
mainly content words including the keywords. This is known and is
routinely used in document processing \cite{ibm}.

We thus conclude that as far as short texts are concerned, the
Zipf's law holds for Chinese characters in the same way as it does
for English words.

{\bf 5.} To check our results on fitting the empiric data for word
frequencies to the Zipf's law we carried out three alternative
tests.

{\bf 5.1} First we applied the Kolmogorov-Smirnov (KS) test to
decide on the fitting quality of the data with the Zipf's law (in
the range $[r_{\rm min}, r_{\rm max}]$).  The test was carried out
both with and without transforming to the logarithmic coordinates
(\ref{kabanyan}) and it fully confirmed our result; see Table
\ref{tab_4}. For a detailed presentation of the KS test results
see Appendix E and Table \ref{tab_8} therein.

{\bf 5.2} It was recently shown that even when the applicability
range $[r_{\rm min}, r_{\rm max}]$ of a power law is known, the
linear least-square method (that we employed above) may not give
accurate estimations for the exponent $\gamma$ of the power law
\cite{Goldstein,Bauke,Newman}.  It was then argued that the method
of Maximum Likelihood Estimation (MLE) is more reliable in this
context. Hence to show that our results are robust, we calculated
$\gamma$ using the MLE method, as suggested in
\cite{Goldstein,Bauke,Newman}. We got that the difference with the
linear least square method is quite small (changes come only at
the third decimal place); see Table \ref{tab_4}.

{\bf 5.3} We also checked whether our results on the power law
exponent $\gamma$ are stable with respect to non-linear fitting
schemes, the ones that do not employ the logarithmic coordinates
(\ref{kabanyan}), but operate directly with the form (\ref{00}).
Again, we find that non-linear fitting schemes (that we carried
out via routines of Mathematica 7) produce very similar results
for $\gamma$; see Table \ref{tab_4}.

One reason for such a good coincidence between our linear fitting
results and alternative tests is that we use a rather strict
criteria ($SS_{\rm err}^*<0.05$ and $R^2>0.995$) for determining
{\it first} the Zipfian range $[r_{\rm min}$, $r_{\rm max}]$ and
then the parameters of the Zipf's law. Another reason is that in
the vicinity of $r_{\rm max}$, the number of different words
having the same frequency is not large (it is smaller than 10).
Hence there are no problems with lack of data points or systematic
biases that can plague the applicability of the least square
method for determination of the exponent $\gamma$.

\begin{table*}
\begin{center}
\tabcolsep0.12in \arrayrulewidth0.5pt
\renewcommand{\arraystretch}{1.3}
\caption{\label{tab_4} Comparison between different methods of
estimating the exponent $\gamma$ of the Zipf's law; see
(\ref{estoup}): LLS (linear least-square), NLS (nonlinear
least-square), MLE (maximum likelihood estimation). We also
present the p-value of the KS test when comparing the empiric word
frequencies in the range $[r_{\rm min}$, $r_{\rm max}]$ with the
Zipf's-law within the linear lest-square method (LLS); for a more
detailed presentation of the KS results see Appendix E. Recall
that the p-values have to be sufficiently larger than $0.1$ for
fitting to be reliable from the viewpoint of KS test. This holds
for the presented data; see Appendix E for details.        }
\vskip 0.3cm
\begin{tabular}{|c|c|c|c|c|}
\hline
Texts & $\gamma$, LLS & $\gamma$, NLS  & $\gamma$, MLE & p-value \\
\hline
TF  & 1.032 & 1.033  & 1.035 & 0.865 \\
TM  & 1.041  &  1.036 &  1.039 & 0.682 \\
AR  &   1.038  &  1.042  &  1.044 & 0.624   \\
DL  &   1.039  &  1.034 &  1.035 &  0.812   \\
AQZ  & 1.03 & 1.028   & 1.027 & 0.587 \\
KLS   &  0.97  &  0.975 & 0.973 & 0.578  \\
CQF  &  0.985  & 0.983  & 0.981 & 0.962 \\
SBZ  &  0.972  & 0.967& 0.973 &  0.796  \\
WJZ  & 0.999  &  0.993  &  0.995 &  0.852  \\
HLJ  & 1.01   & 1.015 &  1.011 &   0.923 \\
\hline
\end{tabular}
\end{center}
\end{table*}

\subsection{Theoretical description of the Zipf's law and hapax legomena}
\label{theoretical_model}

\subsubsection{Assumptions of the model}
\label{3.3.1}

A theoretical description of the Zipf's law that is specifically
applicable to short English texts was recently proposed in
\cite{a}; it is reviewed below. The theory is based on the ideas
of latent semantic analysis and the concept of mental lexicon
\cite{a}.  We shall now briefly remind it to demonstrate that

-- The rank-frequency relation for short Chinese and English texts
can be described by the same theory.

-- The theory allows to extrapolate the Zipf's law to high and low
frequencies (including hapax legomena).

-- It allows to understand the bound $c<0.25$ for the prefactor of
the Zipf's law (since the law does not apply for all frequencies,
$c$ is not fixed from normalization).

-- The theory confirms the intuitive expectation about the
difference between the Zipfian and hapax legomena range: in the
first case the probability of a word is equal to its frequency
(frequent words). In the hapax legomena range, both the
probability and frequency are small and different from each other.


-- In the following section the theory is employed for describing
the rank-frequency relation of Chinese characters outside of the
validity range of the Zipf's law.

Our model for deriving the Zipf's law together with the
description of the hapax legomena makes four assumptions (see
\cite{a} for further details). Below we shall refer to the units
of the text as words; whenever this theory applies for Chinese
texts we shall mean characters instead of words.

$\bullet$ The {\it bag-of-words picture} focusses on the frequency
of the words that occur in a text and neglects their mutual
disposition (i.e. syntactic structure) \cite{madsen}. This is a
natural assumption for a theory describing word frequencies, which
are invariant with respect to an arbitrary permutation of the
words in a text. The latter point was recently verified in
\cite{bernhard}.

Given $n$ different words $\{w_k\}_{k=1}^n$, the joint probability
for $w_k$ to occur $\nu_k\geq 0$ times in a text $T$ is assumed to
be multinomial \BEA \label{multinom} \pi[\nunu
|\ff]=\frac{N!\,\theta_1^{\nu_1}...\theta_n^{\nu_n}}{\nu_1!...\nu_n!},
~~~\nunu = \{\nu_k\}_{k=1}^{n}, ~~ \ff=\{\theta_k\}_{k=1}^{n},
\EEA where $N=\sum_{k=1}^n\nu_k$ is the length of the text
(overall number of words), $\nu_k$ is the number of occurrences of
$w_k$, and $\theta_k$ is the probability of $w_k$.

Hence according to (\ref{multinom}) the text is regarded to be a
sample of word realizations drawn independently with probabilities
$\theta_k$.

The bag-of-words picture is well-known in computational
linguistics \cite{madsen}. But for our purposes it is incomplete,
because it implies that each word has the same probability for
different texts. In contrast, it is well known (and routinely
confirmed by the rank-frequency analysis) that the same words do
{\it not} occur with same frequencies in different texts.

$\bullet$ To improve this point we make $\ff$ a random vector with
a text-dependent density $P(\ff|T)$ (a similar, but stronger
assumption was done in \cite{madsen}). With this assumption the
variation of the word frequencies from one text to another will be
explained by the randomness of the word probabilities.

We now have three random objects: text $T$, probabilities $\ff$
and the occurrence numbers $\nunu$. Since $\ff$ was introduced to
explain the relation of $T$ with $\nunu$, it is natural to assume
that the triple $(T,\ff,\nunu)$ form a Markov chain: the text $T$
influences the observed $\nunu$ only via $\ff$. Then the
probability $p(\nunu|T)$ of $\nunu$ in a given text $T$ reads \BEA
p(\nunu|T)={\int}\d \ff \, \pi[\nunu |\ff]\, P(\ff |T).
\label{bay} \EEA This form of $p(\nunu|T)$ is basic for
probabilistic latent semantic analysis \cite{hof}, a successful
method of computational linguistics. There the density $P(\ff |T)$
of latent variables $\ff$ is determined from the data fitting. We
shall deduce $P(\ff |T)$ theoretically.

$\bullet$ The text-conditioned density $P(\ff |T)$ is generated
from a prior density $P(\ff)$ via conditioning on the ordering of
$\w=\{w_k\}_{k=1}^n$ in $T$: \BEA \label{chua} P(\ff |T)=
P(\ff)\,\chi_T(\ff,\w)\left/ {\int}\d \ff' \, P(\ff')\,
  \chi_T(\ff',\w)\right. .
\EEA Thus if different words of $T$ are ordered as $(w_1,...,w_n)$
with respect to the decreasing frequency of their occurrence in
$T$ (i.e. $w_1$ is more frequent than $w_2$), then
$\chi_T(\ff,\w)=1$ if $\theta_1\geq...\geq\theta_n$, and
$\chi_T(\ff,\w)=0$ otherwise.

$\bullet$ The apriori density of the word probabilities $P(\ff)$
in (\ref{chua}) can be related to the mental lexicon (store of
words) of the author prior to generating a concrete text. For
simplicity, we assume that the probabilities $\theta_k$ are
distributed identically [see \cite{a} for a verification of this
assumption] and the dependence among them is due to
$\sum_{k=1}^n\theta_k=1$ only: \BEA \label{g1} P(\ff)\propto
u(\theta_1)\,...\,u(\theta_n)\,\delta({\sum}_{k=1}^n\theta_k-1),
\EEA where $\delta(x)$ is the delta function and the normalization
ensuring $\int_0^\infty\prod_{k=1}^n \d \theta_k\, P(\ff)=1$ is
omitted.

\subsubsection{Zipf's law}
\label{3.3.2}

It remains to specify the function $u(\theta)$ in (\ref{g1}).
Ref.~\cite{a} reviews in detail the experimentally established
features of the human mental lexicon (see \cite{levelt} in this
context) and deduces from them that the suitable function
$u(\theta)$ is \BEA \label{go3} u(f)=({n}^{-1}{c}+f)^{-2}, \EEA
where $c$ is to be related to the prefactor of the Zipf's law.

Above equations (\ref{multinom}--\ref{go3}) together with the
feature $n^3\gg N\gg 1$ of real texts (where $n$ is the number of
different words, while $N$ is the total number of words in the
text) allow to the final outcome of the theory: the probability
$p_r(\nu|T)$ of the character (or word) with the rank $r$ to
appear $\nu$ times in a text $T$ (with $N$ total characters and
$n$ different characters) \cite{a}: \BEA \label{g00}
p_r(\nu|T)=\frac{N!}{\nu!(N-\nu)!} \phi_r^{\nu}(1-\phi_r)^{N-\nu},
\EEA where the effective probability $\phi_r$ of the character is
found from two equations for two unknowns $\zmu$ and $\phi_r$:
\BEA \label{g4} {r}/{n}= \int_{\phi_r}^\infty \d \f \,
\frac{e^{-\zmu
    \f}}{(c+\f)^2} \left /\int_{0}^\infty \d \f\, \frac{e^{-\zmu
      \f}}{(c+\f)^2}
\right. , \\
\label{g5} \int_{0}^\infty \d \f\, \frac{\f\,e^{-\zmu
\f}}{(c+\f)^2} =\int_{0}^\infty \d \f\, \frac{e^{-\zmu
\f}}{(c+\f)^2}, \EEA where $c$ is a constant that will later on
shown to coincide with the prefactor of the Zipf's law.

For $c\lesssim 0.25$, $c\zmu$ determined from (\ref{g5}) is small
and is found from integration by parts: \BEA \label{simmens}
\zmu\simeq c^{-1}\,e^{-\gamma_{\rm E}-\frac{1+c}{c}}, \EEA where
$\gamma_{\rm E}=0.55117$ is the Euler's constant.  One solves
(\ref{g4}) for $c\zmu\to 0$: \BEA \label{grm} \frac{r}{n}=ce^{-n
\phi_r\zmu}/(c+n \phi_r). \EEA

Recall that according to (\ref{g00}), $\phi_r$ is the probability
for the character (or the word in the English situation) with rank
$r$.  If $\phi_r$ is sufficiently large, $\phi_r N\gg 1$, the
character with rank $r$ appears in the text many times and its
frequency $\nu\equiv f_rN$ is close to its maximally probable
value $\phi_rN$; see (\ref{g00}). Hence the frequency $f_r$ can be
obtained via the probability $\phi_r$.  This is the case in the
Zipfian domain, since according to our empirical results (both for
Chinese and English) $\frac{1}{n}\lesssim f_r$ for $r\leq r_{\rm
max}$, and|upon identifying $\phi_r=f_r$|the above condition
$\phi_r N\gg 1$ is ensured by $N/n\gg 1$; see Tables \ref{tab_1},
\ref{tab_2} and \ref{tab_3}.

Let us return to (\ref{grm}). For $r>r_{\rm min}$,
$\phi_rn\zmu=f_rn\zmu<0.04\ll 1 $; see (\ref{simmens}) and
Figs.~\ref{fig_1} and \ref{fig_2}. We get from (\ref{grm}): \BEA
\label{g6} f_r=c(r^{-1}-n^{-1}). \EEA This is the Zipf's law
generalized by the factor $n^{-1}$ at high ranks $r$. This cut-off
factor ensures faster [than $r^{-1}$] decay of $f_r$ for large
$r$.

Figs.~\ref{fig_1} and \ref{fig_2} shows that (\ref{g6}) reproduces
well the empirical behavior of $f_r$ for $r>r_{\rm min}$. Our
derivation shows that $c$ is the prefactor of the Zipf's law, and
that our assumption on $c\lesssim 0.25$ above (\ref{simmens})
agrees with observations; see Tables \ref{tab_1}, \ref{tab_2} and
\ref{tab_3}.

For given prefactor $c$ and the number of different characters
$n$, (\ref{g4}) predict the Zipfian range $[r_{\rm min},r_{\rm
max}]$ in agreement with empirical results; see Figs.~\ref{fig_1}
and \ref{fig_2}.

For $r<r_{\rm min}$, it is not anymore true that $f_rn\zmu\ll 1$
(though it is still true that $f_r N=\phi_r N\gg 1$). So the
fuller expression (\ref{g4}) is to be used instead of (\ref{grm}).
It reproduces qualitatively the empiric behavior of $f_r$ also for
$r<r_{\rm min}$; see Figs.~\ref{fig_1} and \ref{fig_2}. We do not
expect any better agreement theory and observations for $r<r_{\rm
  min}$, since the behavior of frequencies in this range is irregular
and changes significantly from one text to another.

\subsection{Hapax legomena}
\label{hapoo}

\subsubsection{Hapax legomena as a consequence of the generalized
  Zipf's law}

According to (\ref{g00}), the probability $\phi_r$ is small for
$r\gg r_{\rm max}$ and hence the occurrence number $\nu\equiv
f_rN$ of the character with the rank $r$ is a small integer (e.g.
1 or 2) that cannot be approximated by a continuous function of
$r$; see Figs.~\ref{fig_1} and \ref{fig_2}. In particular, the
reasoning after (\ref{grm}) on the equality between frequency and
probability does not apply, although we see in Figs.~\ref{fig_1}
and \ref{fig_2} that (\ref{g6}) roughly reproduces the trend of
$f_r$ even for $r>r_{\rm
  max}$.

To describe this hapax legomena range, define $r_k$ as the rank,
when $\nu\equiv f_rN$ jumps from integer $k$ to $k+1$ (hence the
number of characters that appear $k+1$ times is $r_{k}-r_{k+1}$).
Since $\phi_r$ reproduces well the trend of $f_r$ even for
$r>r_{\rm max}$, see Fig.~\ref{fig_1}, $r_k$ can be theoretically
predicted from (\ref{g6}) by equating its left-hand-side to $k/N$:
\BEA \hat{r}_k=[\frac{k}{Nc}+\frac{1}{n}]^{-1}, \qquad k=0,1,2,...
\label{hapo} \EEA Eq.~(\ref{hapo}) is exact for $k=0$, and agrees
with $r_k$ for $k\geq 1$; see Table \ref{tab_6}.  We see that a
single formalism describes both the Zipf's law for short texts and
the hapax legomena range. We stress that for describing the hapax
legomena no new parameters are needed; it is based on the same
parameters $N,\, n,\, c$ that appear in the Zipf's law.

\subsubsection{Comparing with previous theories of hapax legomena }
\label{brno}

Several theories were proposed over the years for describing the
hapax legomena range; see \cite{tuldava} for a review. To be
precise, these theories were proposed for rare words (not for rare
Chinese characters), but since the Zipf's law applies to
characters, we expect that these theories will be relevant. We now
compare predictions of the main theories with (\ref{hapo}). The
latter turns out to be superior.

Recall that for obtaining (\ref{hapo}) it is necessary to employ
the generalized (by the factor $n^{-1}$) form (\ref{g6}) of the
Zipf's law. The correction factor is not essential in the proper
Zipfian domain (since it is a pure power law), but is crucial for
obtaining a good agreement with empiric data in the hapax legomena
range; see Figs.~\ref{fig_1} and \ref{fig_2}.  The influence of
this correcting factor can be neglected for $k\gg Nc/n$ in
(\ref{hapo}), where we get
\begin{eqnarray}
  \label{eq:1}
\hat{r}_{k-1}-\hat{r}_{k}\propto \frac{1}{k(k-1)},
\end{eqnarray}
for the number of characters having frequency $k/N$. This
relation, which is a crude particular case of (\ref{hapo}), is
sometimes called the second Zipf's law, or the Lotka's law
\cite{baa,tuldava}. The applicability of (\ref{eq:1}) is however
limited, e.g. it does not apply to the data shown in Table
\ref{tab_6}.

Another approach to frequencies of rare words was proposed in
\cite{kral}; see \cite{tuldava} for a review. Its basic result
(\ref{eq:10}) was recently recovered from a partial maximization
of entropy (random group formation approach) \cite{baek}
\footnote{Ref.~\cite{baek} presented a broad range of
applications,
  but it did not study Chinese characters. We acknowledge one of the
  referees of this work who informed us that such unpublished studies
  do exist: Chinese characters are within the applicability range of
  Ref.~\cite{baek}, as we confirm in Table~\ref{tab_6}. The
  predictions of (\ref{eq:10}) for the AQZ text that we reproduce in
  Table \ref{tab_6} were communicated to us by the referee.  }. It
makes the following prediction for the number $nP(k)$
\footnote{Please
  do not mix up $P(k)$ with the density of character probabilities
  that appear in (\ref{g1}, \ref{go3}). Indeed, $P(k)$ is defined as
  empiric frequency; it has a discrete argument and applies to any
  collection of objects, also the one that was generated by any
  probabilistic mechanism. In contrast, (\ref{g1}, \ref{go3}) amount to
a density of probabities that has continuous argument(s) and
assumes a specific generative model. } of characters that appear
in the text $k$ times (i.e. $P(k)$ is a prediction for
$(r_{k-1}-r_{k})/n$)
\begin{eqnarray}
  \label{eq:10}
P(k)\propto e^{-bk}\, k^{-\gamma}, \qquad 1\leq k\leq f_1N,
\end{eqnarray}
where we omitted the normalization ensuring $\sum_{k=1}^{f_1
  N}P(k)=1$, and where the constants $b>0$ and $\gamma>0$ are
determined from three parameters of the text: the overall number
of characters $N$, the number of different characters $n$ and the
maximal frequency $f_1$ \cite{baek}. Distributions similar to
(\ref{eq:10}) (i.e. exponentially modified power-laws) were
derived from partial maximization of entropy prior to
Ref.~\cite{baek} (e.g. in \cite{dover,vakarin}), but it was
Ref.~\cite{baek} that emphasized their broad applicability.

Note that $P(k)$ in (\ref{eq:10}) does not apply out of the hapax
legomena range, where for all $k$ we must have $P(k)=1/n$.
However, it is expected that for $n\gg 1$ this discrepancy will
not hinder the applicability of (\ref{eq:10}) to $P(k)$ with
sufficiently small values of $k$, i.e. within the hapax legomena
range.

The results predicted by (\ref{eq:10}) are compared with our data
in Table \ref{tab_6}. For clarity, we transform (\ref{eq:10}) to a
prediction $\widetilde{r}_k$ for quantities $r_k$:
\begin{eqnarray}
  \label{eq:2}
  \widetilde{r}_l=n[1-\sum_{k=1}^l P(k)], ~~~ l\geq 1,
\end{eqnarray}
i.e. we go to the cumulative distribution function $\sum_{k=1}^l
P(k)$.

While the predictions of (\ref{eq:2}) are in a certain agreement
with the data, their accuracy is inferior (at least by an order of
magnitude) as compared to predictions of (\ref{hapo}); see Table
\ref{tab_6}. The reason of this inferiority is that though both
(\ref{eq:10}) and (\ref{hapo}) use three input parameters,
(\ref{eq:10}) is not sufficiently specific to the studied text.

Finally, let us turn to the Waring-Herdan approach which predicts
for $nP(k)$ (the number  of characters that appear in the text $k$
times) a version of the Yule's distribution \cite{tuldava}:
\begin{eqnarray}
  \label{eq:3}
  P(k+1)=P(k)\,\frac{a+k-1}{x+k}, \quad k\geq 1,
\end{eqnarray}
where $a$ and $x$ are expressed via three (the same number as in
the previous two approaches) input parameters $N$ (the overall
number of characters), $n$ (the number of distinct characters) and
$nP(1)$ (the number of characters that appear only once)
\cite{tuldava}:
\begin{eqnarray}
  \label{eq:4}
a=\left(\frac{1}{1-P(1)} -P(1)-1\right)^{-1}, ~~~
x=\frac{a}{1-P(1)}.
\end{eqnarray}
Eqs.~(\ref{eq:3}, \ref{eq:4}) are turned to a prediction $r'_k$
for $r_k$. As Table \ref{tab_6_1} shows, these predictions
\footnote{Eq.~(\ref{eq:3}) can viewed as a consequence of the
  Simon's model of text generation. This model does not apply to real
  texts as was recently demonstrated in \cite{bernhard}. Nevertheless
  (\ref{eq:3}) keeps its relevance as a convenient fitting expression;
  see also \cite{tuldava} in this context.}  are also inferior as
compared to those of (\ref{hapo}), especially for $k\geq 5$.

\begin{table*}
\begin{center}
\tabcolsep0.07in \arrayrulewidth0.5pt
\renewcommand{\arraystretch}{1.2}
\caption{\label{tab_6} The hapax legomena range for Chinese
characters
  demonstrated for 4 short Chinese texts. The first and second text
  are in Modern Chinese, other two are in Classic Chinese; see Tables
  \ref{tab_1} and \ref{tab_2}. $r_k$ is defined before (\ref{hapo})
  and is found from empirical data, while $\hat{r}_k$ is calculated
  from (\ref{hapo}); see section \ref{hapoo}. We also present the
  relative error for $\hat{r}_k$ approximating $r_k$.  }  \vskip 0.3cm
\begin{tabular}{|c|c|c|c|c|c|c|c|c|c|c|c|c|}
\hline
Texts & $k$ & 1 & 2 & 3 & 4 & 5 & 6 & 7 & 8 & 9 & 10\\
 \hline \multirow{3}{*}{AQZ}
&$r_k$ & 1097 &857& 702& 595& 522& 461& 414& 370& 339& 311\\
&$\hat{r}_k$ & 1116 &869& 711& 601& 520& 458& 409& 369& 336& 308\\
&$\frac{|\hat{r}_k-r_k|}{r_k}$ & 0.017& 0.014& 0.013& 0.010&
0.0038&
0.0065& 0.012& 0.0027& 0.0088& 0.0096 \\
\hline \multirow{3}{*}{KLS}& $r_k$ & 1405&
1060& 885& 767& 662& 582& 520& 455& 408& 377\\
 &$\hat{r}_k$ & 1428& 1093& 884& 750& 656& 575& 515& 445& 404& 369
\\
&$\frac{|\hat{r}_k-r_k|}{r_k}$ & 0.016& 0.031& 0.0011& 0.022&
0.0091& 0.012& 0.0096& 0.022& 0.0098& 0.021
 \\
\hline \multirow{3}{*}{SBZ}& $r_k$ & 1460&1141& 959& 850& 735
&676& 618& 563& 517& 481
 \\
&$\hat{r}_k$ & 1481& 1168& 980& 848& 740& 656& 599& 553& 497& 488
 \\
&$\frac{|\hat{r}_k-r_k|}{r_k}$ & 0.014& 0.024& 0.022& 0.0024&
0.0068& 0.029& 0.031& 0.018& 0.039& 0.015
\\
\hline \multirow{3}{*}{HLJ}&  $r_k$ & 1302&1045& 872 &756& 669&
 604& 551 &501& 467& 430
\\
&$\hat{r}_k$ & 1327 &1080& 900& 783& 684& 607& 545& 494& 462& 420
\\
&$\frac{|\hat{r}_k-r_k|}{r_k}$ & 0.019& 0.033& 0.032& 0.035&
0.022&
 0.0049& 0.011& 0.014& 0.011& 0.023
\\
\hline
\end{tabular}
\end{center}
\end{table*}

\begin{table*}
\begin{center}
\tabcolsep0.07in \arrayrulewidth0.5pt
\renewcommand{\arraystretch}{1.2}
\caption{\label{tab_6_1} The hapax legomena range for 2 Chinese
texts;
  see Table \ref{tab_1} and cf. with Table \ref{tab_6}. We compare the
  relative errors for, respectively, $\hat{r}_k$ (given by
  (\ref{hapo})) $\widetilde{r}_k$ and $r'_k$ in approximating the data $r_k$;
  see section \ref{brno}. Here $\widetilde{r}_k$ is defined by
  (\ref{eq:2}, \ref{eq:10}), and $r'_k$  is the prediction made by
  (\ref{eq:3}, \ref{eq:4}). For AQZ the parameters in (\ref{eq:10})
  are $\gamma=1.443$ and $b=0.0049$. For KLS: $\gamma=1.574$ and
  $b=0.0033$. It is seen that the relative error provided by
  $\hat{r}_k$ is always smaller; the only exclusion is the case $k=2$
  of the KLS text. Recall that $r'_1=r_1$ by definition. }
\vskip 0.3cm
\begin{tabular}{|c|c|c|c|c|c|c|c|c|c|c|c|c|}
  \hline
  Texts & $k$ & 1 & 2 & 3 & 4 & 5 & 6 & 7 & 8 & 9 & 10\\
  \hline \multirow{3}{*}{AQZ}
  &${|\widetilde{r}_k-r_k|}/\,{r_k}$ & 0.141 & 0.161 & 0.153 & 0.138 &
  0.129 & 0.112 & 0.097 & 0.069 & 0.057 & 0.039 \\
  &${|r'_k-r_k|}/\,{r_k}$ & 0& 0.025&0.048& 0.071& 0.102& 0.121& 0.141& 0.146& 0.163& 0.174 \\
  &${|\hat{r}_k-r_k|}/\,{r_k}$ & 0.017& 0.014& 0.013& 0.010& 0.0038&
  0.0065& 0.012& 0.0027& 0.0088& 0.0096\\
  \hline \multirow{3}{*}{KLS}
  &${|\widetilde{r}_k-r_k|}/\,{r_k}$ & 0.194& 0.221& 0.245& 0.267&
  0.259& 0.250& 0.240& 0.206& 0.183& 0.179 \\
  &${|r'_k-r_k|}/\,{r_k}$ & 0& 0.0087& 0.063& 0.114& 0.137& 0.157& 0.176& 0.168& 0.170& 0.190 \\
  &${|\hat{r}_k-r_k|}/\,{r_k}$ & 0.016& 0.031& 0.0011& 0.022&
  0.0091& 0.012& 0.0096& 0.022& 0.0098& 0.021
  \\
  \hline
\end{tabular}
\end{center}
\end{table*}

\subsection{Summary}

It is to be concluded from this section that|as far as the
applicability of the Zipf's law to short texts is concerned|the
Chinese characters behave similarly to English words. In
particular, both situations can be adequately described by the
same theory. In particular, the hapax legomena range of short
texts is described via the generalized Zipf's law.

We should like to stress again why the consideration of short
texts is important. One can argue that|at least for the sake of
rank-frequency relations|long texts are just mixtures (joinings)
of shorter, thematically homogeneous pieces (this premise is fully
confirmed below). Hence the task of studying rank-frequency
relations separates into two parts: first understanding short
texts, and then long ones. We now move to the second part.

\section{Rank-frequency relation for long texts and mixtures of texts}
\label{long}

\subsection{Mixing English texts}

When mixing (joining)\footnote{Upon joining two texts (A and B),
the
  word frequencies get mixed: $f_k({\rm A\& B})=\frac{N_{\rm
      A}}{N_{\rm A}+N_{\rm B}}f_k({\rm A})+ \frac{N_{\rm B}}{N_{\rm
      A}+N_{\rm B}}f_k({\rm B})$, where $N_{\rm A}$ and $f_k({\rm A})$
  are, respectively, the total number of words and the frequency of
  word $k$ in the text A.  } different English texts the validity
range of the Zipf's law increases due to acquiring more higher
rank words, i.e.  $r_{\rm min}$ stays approximately fixed, while
$r_{\rm
  max}$ increases; see Table \ref{tab_3}. The overall precision of the
Zipf's law also increases upon mixing, as Table \ref{tab_3} shows.

\begin{figure}
\begin{center}
\resizebox{0.95\columnwidth}{!}{%
 \includegraphics{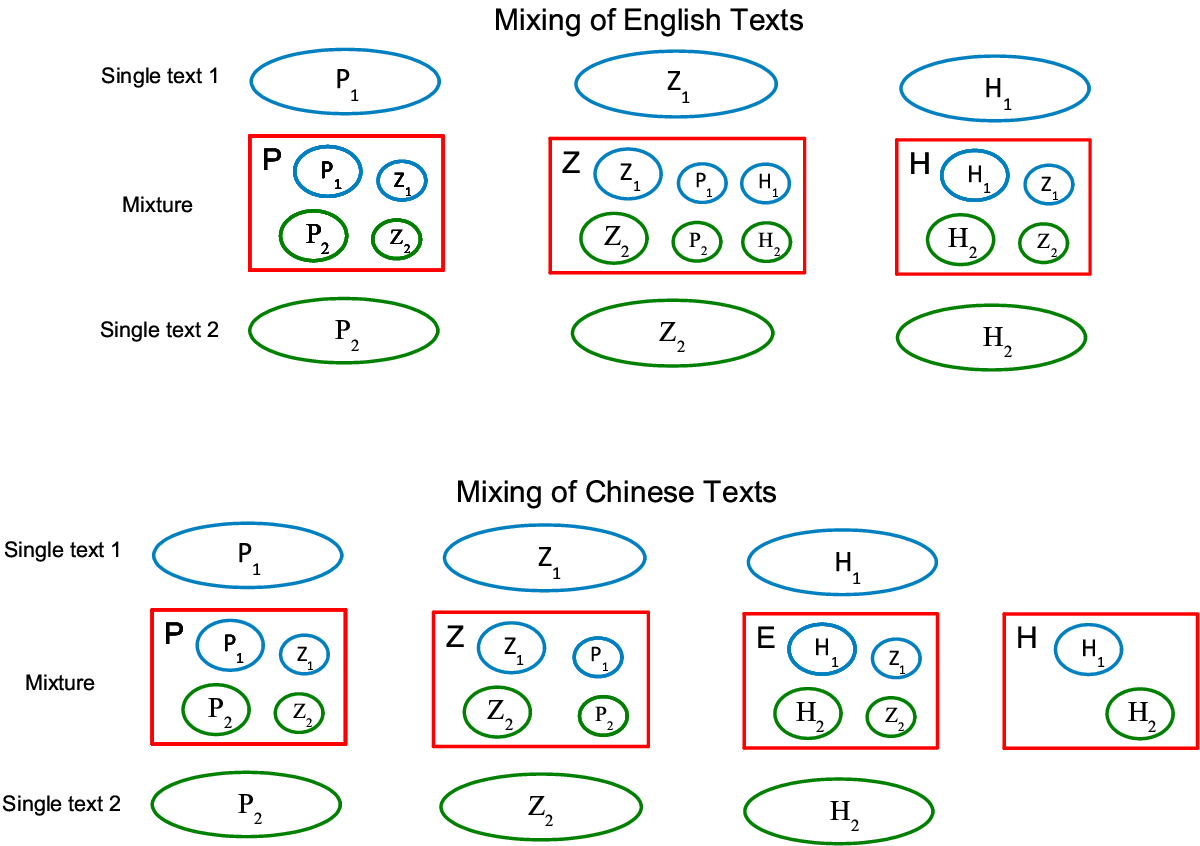} } \vskip 0.2cm
\caption{\label{fig_3} Schematic representation of various ranges
under mixing (joining) two English (upper figure) and two Chinese
(lower figure) texts. ${\rm P}_k$, ${\rm Z}_k$ and ${\rm H}_k$
mean, respectively, the pre-Zipfian, Zipfian and hapax legomena
ranges of the text $k$ ($k=1,2$). ${\rm P}$, ${\rm Z}$ and ${\rm
H}$ mean the corresponding ranges for the mixture of texts $1$ and
$2$. E means the exponential-like range that emerges upon mixing
of two Chinese texts. For each range of the mixture we show
schematically contributions from various ranges of the separate
texts. The relative importance of each contribution is
conventionally represented by different magnitudes of the circles.
}
\end{center}
\end{figure}

The rough picture of the evolution of the rank-frequency relation
under mixing two texts is summarized as follows; see Table
\ref{tab_3} and Fig.~\ref{fig_3} for a schematic illustration. The
majority of the words in the Zipfian range of the mixture (e.g. AR
\& TM) come from the Zipfian ranges of the separate texts. In
particular, all the words that appear in the Zipfian ranges of the
separate words do appear as well in the Zipfian range of the
mixture (e.g. the Zipfian ranges of AR and TM have 130 common
words). There are also relatively smaller contributions to the
Zipfian range of the mixture from the pre-Zipfian and hapax
legomena range of separate texts: note from Table \ref{tab_3} that
the Zipfian range of the mixture AR \& TM is 82 words larger than
the sum of two separate Zipfian ranges, which is (307 + 290) minus
130 common words.

Some of the words that appear only in the Zipfian range of one of
separate texts will appear in the hapax legomena range of the
mixture; other words move from the pre-Zipfian range of separate
texts to the Zipfian range of the mixture. But these are
relatively minor effects: the rough effect of mixing is visualized
by saying that the Zipfian ranges of both texts combine to become
a larger Zipfian range of the mixture and acquire additional words
from other ranges of the separate texts; see Fig.~\ref{fig_3}.
Note that the keywords of separate words stay in the Zipfian range
of the mixture, e.g. after joining all four above texts, the
keywords of each text are still in the Zipfian range (which now
contains almost 900 words); see Table \ref{tab_3}.

The results on the behavior of the Zipf's law under mixing are
new, but their overall message|the validity of the Zipf's law
improves upon mixing|is expected, since it is known that the
Zipf's law holds not only for short but also for long English
texts and for frequency dictionaries (huge mixtures of various
texts) \cite{wyllys,book,baa,li}.

\subsection{Mixing Chinese texts}

\subsubsection{Stability of the Zipfian range}

The situation for Chinese texts is different. Upon mixing two
Chinese texts the validity range of the Zipf's law increases, but
much slower as compared to English texts; see Tables \ref{tab_1}
and \ref{tab_2}. The validity ranges of the separate texts do not
combine (in the above sense of English texts). Though the common
words in the Zipfian ranges of separate texts do appear in the
Zipfian range of the mixture, a sizable amount of those words that
appeared in the Zipfian range of only one text do not show up in
the Zipfian range of the mixture \footnote{ As an example, let us
consider in detail the mixing
  of two Chinese texts SBZ and CQF; see Table \ref{tab_2}.  The
  Zipfian ranges of CQF and SBZ contain, respectively, 306 and 319
  characters. Among them 133 characters are common.  The balance of
  the characters upon mixing is calculated as follows: 306 (from the
  Zipfian range of CQF) + 319 (from the Zipfian range of SBZ) - 133
  (common characters) - 50 (characters from the Zipfian range of CQF
  that do not appear in the Zipfian range of CQF \& SBZ) - 54
  (characters from the Zipfian range of SBZ that do not appear in the
  Zipfian range of CQF+SBZ) +27 (characters that enter to the Zipfian
  range CQF \& SBZ from the pre-Zipfian ranges of CQF or SBZ)= 415
  (characters in the Zipfian range of CQF \& SBZ). }.

Importantly, the overall frequency of the Zipfian domain for very
different Chinese texts (mixtures, long texts) is approximately
the same and amounts to $\simeq 0.4$; see Tables \ref{tab_1} and
\ref{tab_2}. In contrast, for English texts this overall frequency
grows with the number of different words in the text; see Table
\ref{tab_3}. This is consistent with the fact that for English
texts the Zipfian range increases upon mixing.

\begin{figure}
\begin{center}
\resizebox{0.82\columnwidth}{!}{%
 \includegraphics{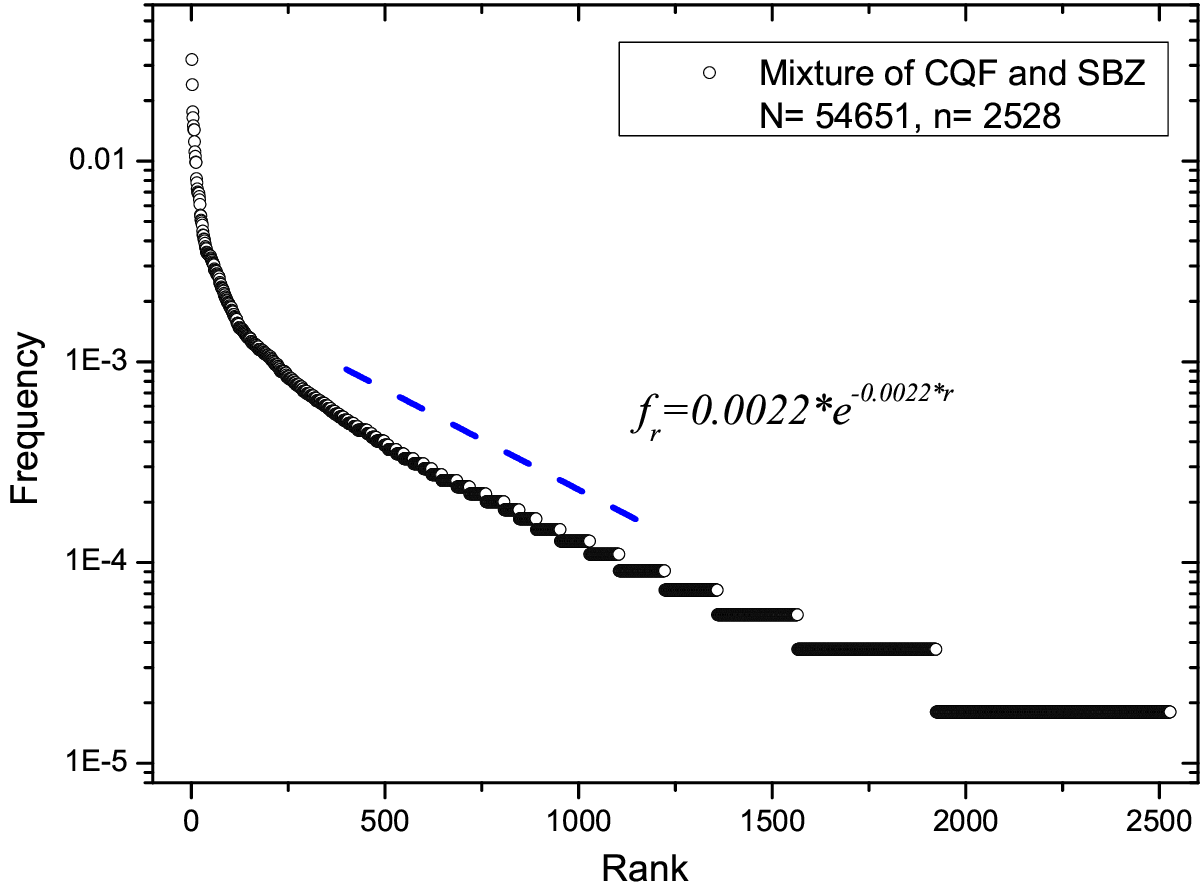} }
\caption{\label{fig_4} (Color online) Rank frequency distribution
for the mixture of
  CQF and SBZ.; see Tables \ref{tab_1} and \ref{tab_2} and Appendix
  C. The scale of the frequency is chosen such that the
  exponential-like range of the rank-frequency relation for $r>500$ is
  made visible. For comparison, the dashed blue line shows a curve
  $f_r=0.0022 e^{-0.0022 r}$. For the present example, the
  exponential-like range is essentially mixed with hapax legomena,
  since for frequencies $f_r$ with $r>r_{\rm max}$ the number of
  different words having this frequency is larger than 10. Recall that
  the Zipf's law holds for $r_{\rm min}<r<r_{\rm max}$; see Tables
  \ref{tab_1} and \ref{tab_2}. }
\end{center}
\end{figure}

\subsubsection{Emergence of the exponential-like range}
\label{expo}

The majority of characters that appear in the Zipfian range of
separate texts, but do not appear in the Zipfian range of the
mixture, moves to the hapax legomena range of the mixture. Then,
for larger mixtures and longer texts, a new, exponential-like
range of the rank-frequency relation emerges from within the hapax
legomena range.

To illustrate the emergence of the exponential-like range let us
start with Fig.~\ref{fig_4}. Here there are only two short texts
mixed and hence the exponential-like range cannot be reliably
distinguished from the hapax legomena \footnote{Recall in this
context that in the hapax
  legomena range many characters have the same frequency, hence no
  smooth rank-frequency relation is reliable.}: for all frequencies
with the ranks $r>r_{\rm max}$ (i.e. for all frequencies beyond
the Zipfian range), the number of different characters having
exactly the same frequency is larger than 10. (We conventionally
take this number as a borderline of the hapax legomena.) However,
the trace of the exponential-like range is seen even within the
hapax legomena; see Fig.~\ref{fig_3}.

\begin{table*}
\begin{center}
\tabcolsep0.08in \arrayrulewidth0.5pt
\renewcommand{\arraystretch}{1.3}
\caption{\label{tab_5} Parameters of the exponential-like range
(lower and upper ranks and the overall frequency) for few long
Chinese texts; see also Tables \ref{tab_1} and \ref{tab_2}. Here
$n$ is the number of different characters. Recall that the lowest
rank of the exponential-like range is $r_{\rm max}+1$, where
$r_{\rm max}$ is the upper rank of the Zipfian range. The highest
rank of the exponential-like range was denoted as $r_b$; see
Tables \ref{tab_1} and \ref{tab_2}. }  \vskip 0.3cm
\begin{tabular}{|c|c|c|c|}
\hline
Texts & $n$ & Rank range & Overall frequency \\
\hline
PFSJ  & 3820 & 584--1437  & 0.12816 \\
\hline
SHZ   & 4376     & 591--1618   & 0.14317 \\
\hline
SJ   &  4932     & 536--1336   & 0.12887 \\
\hline
14 texts & 5018  & 626--1223   & 0.12291 \\
\hline
\end{tabular}
\end{center}
\end{table*}

\begin{figure}
\begin{center}
\resizebox{0.85\columnwidth}{!}{%
 \includegraphics{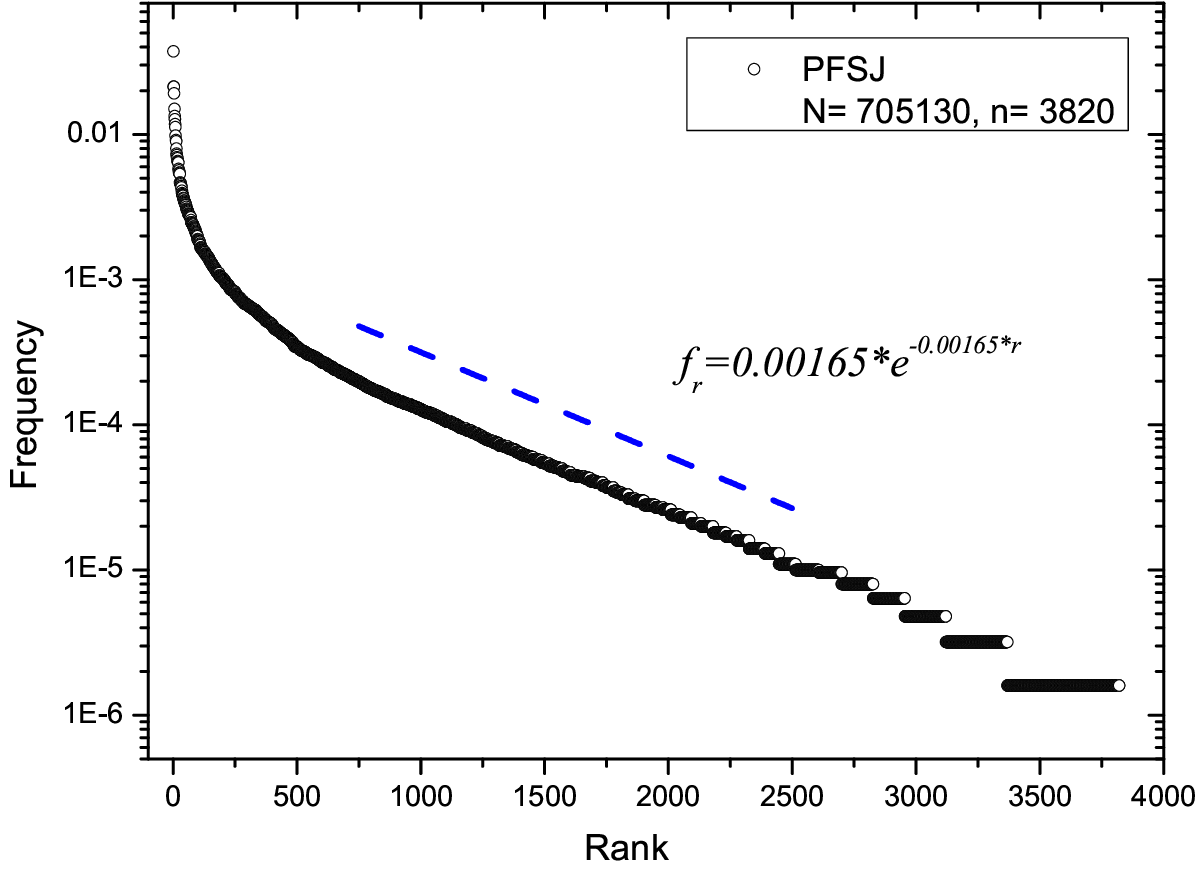} }
\caption{\label{fig_5} (Color online) Rank frequency distribution
of the long modern Chinese text PFSJ. The exponential behavior
$f_r\propto e^{-0.00165 r}$ of frequency $f_r$ is visible for
$r>500$. For comparison, the dashed blue line shows a curve
$f_r=0.00165 e^{-0.00165 r}$. The boundary between the
exponential-like range and hapax legomena can be defined as the
rank $r_b$, where the number of words having the same frequency
$f_{r_b}$ is equal to $10$. For the present example $r_b=1437$.
The Zipf's law holds for ranks $r_{\rm min}<r<r_{\rm max}$, where
$r_{\rm max}=583$, $r_{\rm min}=67$; see Table \ref{tab_1}. }
\end{center}
\end{figure}

For bigger mixtures or longer texts, the exponential-like range
clearly differentiates from the hapax legomena. In this context,
we define $r_b$ as the borderline rank of the hapax legomena: for
$r>r_b$, the number of characters having the frequency $f_{r_b}$
is larger than 10.  Then the exponential-like range \BEA f_r = a
e^{-br}~~ {\rm with} ~~ a<b, \EEA exists for the ranks $r_{\rm
max}< r\lesssim r_b$ (provided that $r_{\rm max}$ is sufficiently
larger than $r_b$); see Table \ref{tab_5}.  Put differently, the
exponential-like range exists from ranks larger than the upper
rank $r_{\rm max}$ of the Zipfian range till the ranks, where the
hapax legomenon starts. Tables \ref{tab_1}, \ref{tab_2},
\ref{tab_5} and Fig.~\ref{fig_5} show that the exponential-like
range is not only sizable by itself, but (for sufficiently long
texts or sufficiently big mixtures) it is also bigger than the
Zipfian range. This, of course, does not mean that the Zipfian
range becomes less important, since, as we saw above, it carries
out nearly 40 \% of the overall frequency; see Tables \ref{tab_1}
and \ref{tab_2}. The exponential-like range also carries out
non-negligible frequency, though it is few times smaller than that
of the Zipfian and pre-Zipfian ranges; see Tables \ref{tab_1},
\ref{tab_2} and \ref{tab_5}.

Finally, we would like to stress that we considered various
Chinese texts written with simplified or traditional characters,
with Modern Chinese or different versions of Classic Chinese; see
Tables \ref{tab_1}, \ref{tab_2} and Appendix C.  As far as the
rank-frequency relations are concerned, all these texts
demonstrate the same features showing that the peculiarities of
these relations are based on certain very basic features of
Chinese characters. They do not depend on specific details of
texts.

\subsection{Theoretical description of the exponential-like regime}
 \label{4.3}

\begin{figure}
\begin{center}
\resizebox{0.85\columnwidth}{!}{%
 \includegraphics{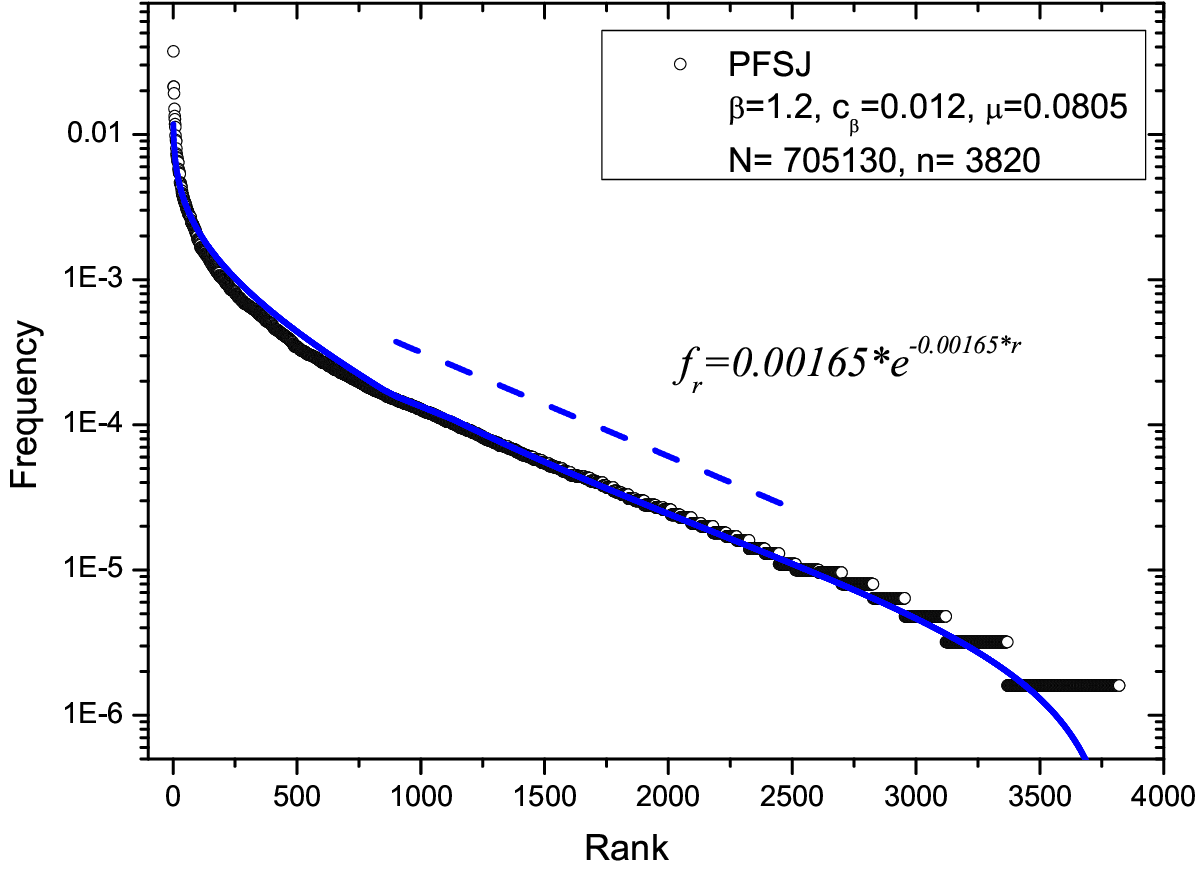}  }
\caption{\label{fig_6} (Color online) The rank-frequency relation
$f(r)$ for characters from the text PFSJ; see Table \ref{tab_1}.
Blue line denotes the numerical solution of (\ref{gk4}, \ref{gk5})
at the indicated parameters $\beta$ and $c_\beta$. The dashed blue
line indicates at the exponential-like regime.  }
\end{center}
\end{figure}

\begin{figure}
\begin{center}
\resizebox{0.85\columnwidth}{!}{%
 \includegraphics{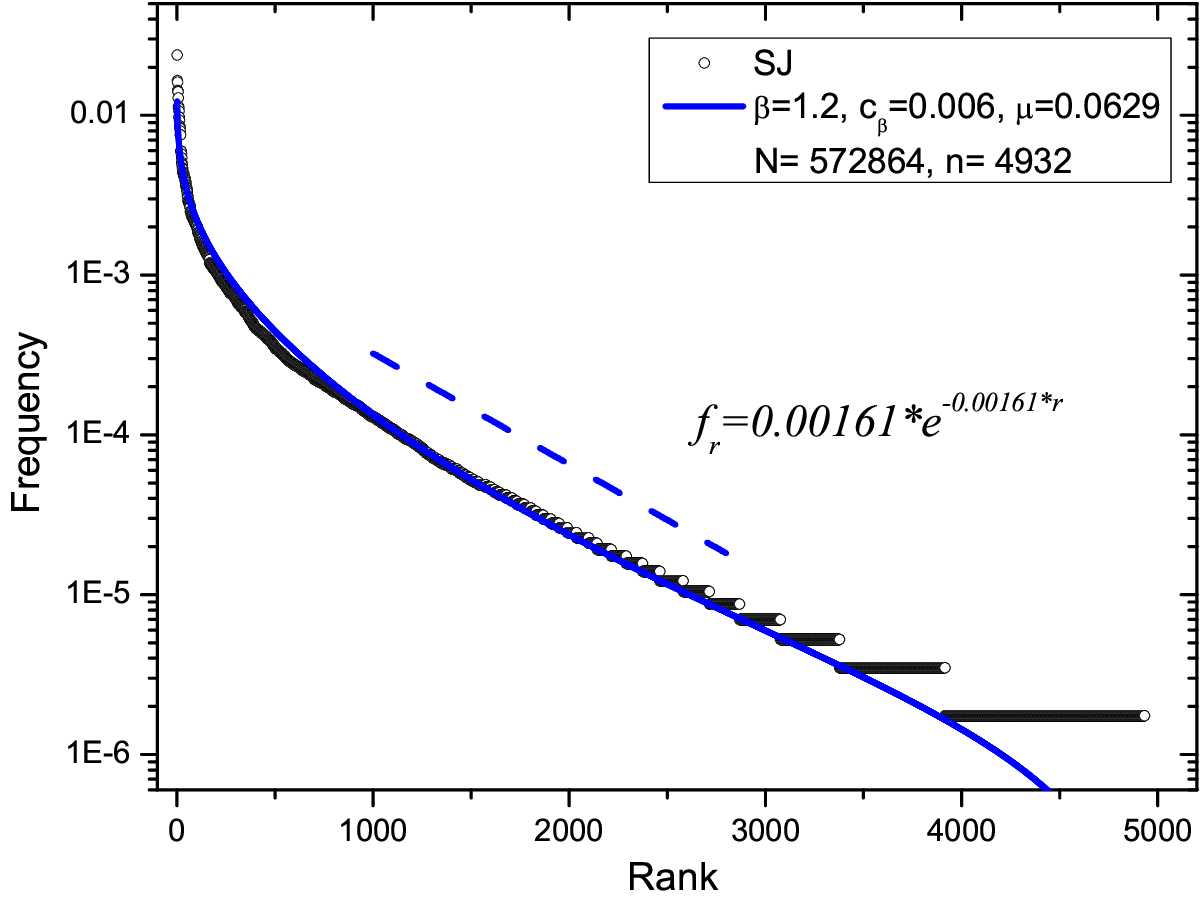}  }
\caption{\label{fig_7} (Color online) The rank-frequency relation
$f(r)$ for characters from the text SJ; see Table \ref{tab_2}. For
parameters and notations see Fig.~6.  }
\end{center}
\end{figure}

 Now we search for a theoretical description for the exponential like
 regime of the rank-frequency relation of Chinese characters. This
 description will simultaneously account for the hapax legomena range
 (rare words) of long Chinese texts.

 We proceed with the theory outlined in section \ref{3.3.1} and
 \ref{3.3.2}. There we saw that the Zipf's law results from the choice
 (\ref{go3}) of the prior density for word probabilities $\ff$. Now we
 need to generalize (\ref{go3}). Recall that the choice of prior
 densities is is the main problem of the Bayesian statistics
 \cite{jaynes,jaeger} \footnote{We stress that (for a continuous event
   space) this problem is not solved by the maximum entropy method. In
   contrast, this method itself does need the prior density as one of
   its inputs \cite{jaynes}. }. One way to approach this problem is to
 look for a natural group in the space of events (e.g. the translation
 group if the event space is the real line) and then define the
 non-informative prior density as the one which is invariant with
 respect to the group \cite{jaynes,jaeger}. Our event space is the
 simplex $\ff\in{\cal S}_n$: the set of $n$ non-negative numbers (word
 probabilities) that sum to one. The natural group on the simplex is
 the multiplicative group \cite{jaynes} (in a sense this is the only
 group that preserves probability relations \cite{jaeger}), and the
 corresponding non-informative density is the Haldane's prior
 \cite{haldane,jaynes,jaeger} that is given by (\ref{g1}) under
 \BEA \label{gogo3} u(f)=({n}^{-1}c_1+f)^{-1}, ~~~ c_1\to 0. \EEA The
 formal Haldane's prior is recovered from (\ref{gogo3}) under
 $c_1\equiv 0$; a small but finite constant $c_1$ is necessary for
 making the density normalizable.

Note that the prior density (\ref{go3}) which supports the Zipf's
law is far from being non-informative. This is natural, because it
relates to a definite organization of the mental lexicon \cite{a}.

Now the exponential-like regime of the rank-frequency relation can
be deduced via a prior density that is intermediate between the
Zipfian prior (\ref{go3}) and the non-informative, Haldane's prior
(\ref{gogo3}) \BEA \label{gogo4}
u(f)=({n}^{-1}c_\beta+f)^{-\beta}, ~~~1<\beta< 2, \EEA where
$\beta$ and $c_\beta>0$ are to be determined from the
data-fitting. Now we can still use (\ref{g00}) but instead of
(\ref{g4}, \ref{g5}) we get the following implicit relations for
the smooth part of the rank-frequency relation $f_r$ \BEA
\label{gk4} {r}/{n}= \int_{f_r}^\infty \d \f \, \frac{e^{-\zmu
    \f}}{(c_\beta+\f)^\beta} \left /\int_{0}^\infty \d \f\,
  \frac{e^{-\zmu \f}}{(c_\beta+\f)^\beta}
\right. , \\
\label{gk5} \int_{0}^\infty \d \f\, \frac{\f\,e^{-\zmu
    \f}}{(c_\beta+\f)^\beta} =\int_{0}^\infty \d \f\, \frac{e^{-\zmu
    \f}}{(c_\beta+\f)^\beta}.
\EEA

Figs.~\ref{fig_6} and \ref{fig_7} compare these analytical
predictions with data. The fit is seen to be good under parameters
$\beta$ and $c_\beta$ that are not very far from (\ref{gogo3}).
The fact that prior densities close to the non-informative
(Haldane's) prior generate an exponential-like shape for the
rank-frequency relations is intuitive, since such a shape means
that a relatively small group of words carries out the major part
of frequency.

As confirmed by Figs.~\ref{fig_6} and \ref{fig_7}, predictions of
(\ref{gk4}, \ref{gk5}) that describe the exponential-like regime
are not applicable for the Zipfian range.

Importantly, (\ref{gk4}, \ref{gk5}) allow to describe the hapax
legomena range of long Chinese texts. Following section
\ref{hapoo}, we equate the solution $f_r$ of (\ref{gk4},
\ref{gk5}) to $k/N$ and determine from this $r=\hat{r}_k$: the
rank in the hapax legomena range, where the frequency jumps from
$k/N$ to $(k+1)/N$. Now $\hat{r}_k$ agrees well with the empiric
data for the hapax legomena range of long Chinese texts, and the
agreement is better than for the approach based on (\ref{eq:10},
\ref{eq:2}); see Table \ref{tab_6_2} and cf. with Table
\ref{tab_6_1}.

Note that the range of rare words (hapax legomena) relates to that
part of the rank-frequency relation which is closest to it, i.e.
for long Chinese texts it relates to the exponential-like regime
and not to the Zipfian regime.

Though suggestive, the above theoretical results are still
preliminary. The full theory of the rank-frequency relations for
Chinese characters should really {\it explain} how specifically a
non-Zipfian relations result from mixing texts that separately
hold the Zipf's law.

\begin{table*}
\begin{center}
\tabcolsep0.07in \arrayrulewidth0.5pt
\renewcommand{\arraystretch}{1.2}
\caption{\label{tab_6_2} The hapax legomena range for 2 Chinese
texts;
  see Tables \ref{tab_1} and \ref{tab_2}; cf. with Table \ref{tab_6}. We compare the
  relative errors for, respectively, $\hat{r}_k$ (given by
  (\ref{hapo})) and $\widetilde{r}_k$ in approximating the data $r_k$;
  see section \ref{brno}. Here $\widetilde{r}_k$ is defined by
  (\ref{eq:2}, \ref{eq:10}). For PFSJ the parameters in (\ref{eq:10})
  are $\gamma=1.302$ and $b=0.00013$. For SJ: $\gamma=1.299$ and
  $b=0.00026$. It is seen that the relative error provided by
  $\hat{r}_k$ is always significantly smaller. }
\vskip 0.3cm
\begin{tabular}{|c|c|c|c|c|c|c|c|c|c|c|c|c|}
  \hline
  Texts & $k$ & 1 & 2 & 3 & 4 & 5 & 6 & 7 & 8 & 9 & 10\\
  \hline \multirow{2}{*}{PFSJ}
  &${|\widetilde{r}_k-r_k|}/\,{r_k}$ & 0.179 & 0.251 & 0.294 & 0.324 &
  0.340 & 0.356 & 0.365 & 0.376 & 0.387 & 0.392 \\
  &${|\hat{r}_k-r_k|}/\,{r_k}$ & 0.020& 0.017& 0.008& 0.001& 0.002&
  0.008& 0.008& 0.011& 0.009& 0.011\\
  \hline \multirow{2}{*}{SJ}
  &${|\widetilde{r}_k-r_k|}/\,{r_k}$ & 0.093& 0.115& 0.136& 0.153&
  0.167& 0.176& 0.181& 0.189& 0.195& 0.198 \\
  &${|\hat{r}_k-r_k|}/\,{r_k}$ & 0.020& 0.018& 0.011& 0.001&
  0.007& 0.013& 0.0011& 0.012& 0.008& 0.004
  \\
  \hline
\end{tabular}
\end{center}
\end{table*}

\section{Discussion}

\subsection{Summary of results}

~~~~{\bf 1.} As implied by the rank-frequency relation for
characters, short Chinese texts demonstrate the same Zipf's
law|together with its generalization to high and low frequencies
(rare words)|as short English texts; see section III. Assuming
that authors write mainly relatively short texts (longer texts are
obtained by mixing shorter ones), this similarity implies that
Chinese characters play the same role as English words; see
Footnote \ref{neo} in this context. Recall from section II that
{\it a priori} there are several factors which prevent a direct
analogy between words and characters.

{\bf 2.} As compared to English, there are two novelties of the
rank-frequency relation of Chinese characters in long texts.

{\bf 2.1} The overall frequency of the Zipfian range (the range of
middle ranks, where the Zipf's law holds) stabilizes at $\simeq
0.4$. This holds for all texts we studied (written in different
epochs, genres with different types of characters; see Tables
\ref{tab_1}, \ref{tab_2} and Appendix C). A similar stabilization
effect holds as well for the overall frequency of the pre-Zipfian
range for both English and Chinese texts; see Tables \ref{tab_1},
\ref{tab_2} and \ref{tab_3}.

{\bf 2.2} There is a range with an exponential-like rank-frequency
relation. It emerges for relatively longer texts from within the
range of rare words (hapax legomena). The range of ranks, where
the exponential-like regime holds, is larger than that of the
Zipf's law. But its overall frequency is few times smaller; see
Tables \ref{tab_1}, \ref{tab_2} and \ref{tab_5}.

Both these results are absent for English texts; there the overall
frequency of the Zipfian range grows with the length of the text,
while there is no exponential-like regime: the Zipfian range end
with the hapax legomena; see Table \ref{tab_3} and
Fig.~\ref{fig_2}.

The results {\bf 2.1} and {\bf 2.2} imply that long Chinese texts
do have a hierarchic structure: there is a group of characters
that hold the Zipf's law with nearly universal overall frequency
equal to $\simeq 0.4$, and yet another group of relatively less
frequent characters that display the exponential-like range of the
rank-frequency relation.

\subsection{Interpretation of results}

Chinese characters differ from English words, since only long
Chinese texts have the above hierarchic structure.  The underlying
reason of the hierarchic structure is to be sought via the
linguistic differences between Chinese characters and English
words, as we outlined in section II. In particular, the features
{\bf 4, 6, 7} discussed in section II can mean that certain
homographic content characters play multiple role in different
parts of a long Chinese text. They are hence distinguished and
appear in the Zipfian range of the long text with (approximately)
stable overall frequency $\simeq 0.4$. Since this frequency is
sizable, and since the range of ranks carried out by the Zipf's
law is relatively small, there is a relatively large range of
ranks that has to have a relatively small overall frequency; cf.
Tables \ref{tab_1}, \ref{tab_2} with Table \ref{tab_5}. It is then
natural that in this range there emerges an exponential-like
regime that is related with a faster (compared to a power law)
decay of frequency versus rank.

Recall that the stabilization holds as well for the overall
frequency of the pre-Zipfian domain both for English and Chinese
texts. The explanation of this effect is similar to that given
above (but to some extent is also more transparent): the
pre-Zipfian range contains mostly function characters, which are
not specific and used in different texts. Hence upon mixing the
pre-Zipfian range has a stable overall frequency.

The above explanation for the coexistence of the Zipfian and
exponential-like range suggests that there is a relation between
the characters that appear in the Zipfian range of long texts and
homography. As a preliminary support for this hypothesis, we
considered the following construction.  Assuming that a mixture is
formed from separate texts $T_1,..., T_k$, we looked at characters
that appear in the Zipfian ranges of all the separate texts
$T_1,..., T_k$; see Table \ref{tab_2} for examples. This
guarantees that these characters appear in the Zipfian range of
the mixture. Then we estimated (via an explanatory dictionary of
Chinese characters) the average number of different meanings for
these characters. This average number appeared to be around $8$,
which is larger than the average number of meanings for an
arbitrary Chinese character (i.e. when the averaging is taken over
all characters in the dictionary) that is known to be not larger
than 2 \cite{obukhova}.

We should like to stress however that the above connection between
the uncovered hierarchic structure and the number of meanings is
preliminary, since we currently lack a reliable scheme of relating
the rank-frequency relation of a given text to its semantic
features; for a recent review on the (lexical) meaning and its
disambiguation within machine learning algorithms see \cite{book}.

\subsection{Conclusion}

The above discussion makes clear that a theory for studying the
rank-frequency relation of a long text, as it emerges from mixing
of different short texts, is currently lacking. Such a theory was
not urgently needed for English texts, because there the
(generalized) Zipf's law (\ref{g6}) describes well both long and
short texts. But the example of Chinese characters clearly shows
that the changes of the rank-frequency relation under mixing are
essential. Hence the theory of the effect is needed.

Finally, one of main open questions is whether the uncovered
hierarchical structure is really specific for Chinese characters,
or it will show up as well for English texts, but on the level of
the rank-frequency relation for morphemes and not the words.
Factorizing English words into proper morphemes is not
straightforward, but still possible.

\section*{Acknowledgments}

This work is supported by the Region des Pays de la Loire under
the Grant 2010-11967, and by the National Natural Science
Foundation of China (Grant Nos. 10975057, 10635020, 10647125 and
10975062), the Programme of Introducing Talents of Discipline to
Universities under Grant No. B08033, and the PHC CAI YUAN PEI
Programme (LIU JIN OU [2010] No. 6050) under Grant No. 2010008104.

\section*{Appendix A: Glossary}
\label{gloss}

\ \ \ \ $\bullet$ \underline{Classic Chinese}: ({\it w\'en y\'an})
written language employed in China till the early XX (20th)
century. It lost its official status and was changed to Modern
Chinese since the May Fourth Movement in 1919. The Modern Chinese
keeps many elements of Classic Chinese. As compared to the Modern
Chinese, the Classic Chinese has the following peculiarities (1)
It is more lapidary: texts contain almost two times smaller amount
of characters, since the Classic Chinese is dominated by
one-character words. (2) It lacks punctuation signs and affixes.
(3) It relies more on the context. (4) It frequently omits
grammatical subjects.

$\bullet$ \underline{Content} word (character): a word that has a
meaning which can be explained independently from any sentence in
which the word may occur.  Content words are said to have lexical
meaning, rather than indicating a syntactic (grammatical)
function, as a function word does.

$\bullet$ \underline{Empty} Chinese characters|e.g. ``¼¸" {\it
(j\v\i)} or ``ÒÑ" {\it (y\v\i)} |serve for establishing numerals
for nouns, aspects for verbs {\it etc}. In contrast, to
\underline{function} characters, they cannot be used alone, i.e.
they are fully bound.

$\bullet$ \underline{Frequency dictionary}: collects words used in
some activity (e.g. in exact science, or daily newspapers {\it
etc}) and orders those words according to the frequency of usage.
Frequency dictionaries can be viewed as big mixtures of different
texts.

$\bullet$ \underline{Function} word (character): is a word that
has little lexical meaning or have ambiguous meaning, but instead
serves to express grammatical relationships with other words
within a sentence, or specify the attitude or mood of the speaker.
Such words are said to have a grammatical meaning mainly, e.g.
{\it the} or {\it and}.

$\bullet$ \underline{Hapax legomena}: literally means the set of
words (characters) that appeared only once in a text. We employ
this term in a broader sense: the set of words (characters) that
appear in a text only few times. Operationally, this set is
characterized by the fact that sufficiently many words
(characters) have the same frequency. Texts written by human
subjects contains a sizable hapax legomena. This is a non-trivial
fact, since it is not difficult to imagine an artificial text (or
purposefully modified natural text) that will not contain at all
words that appear only once.

$\bullet$ \underline{Homophones}: two different words that are
pronounced in the same way, but may be written differently (and
hence normally have different meaning), e.g. {\it rain} and {\it
reign}.

$\bullet$ \underline{Homographs}: two different words (or
characters) that are written in the same way, but may be
pronounced differently, e.g. {\it shower} [precipitation] and {\it
shower} [the one who shows]. This example is a proper homograph,
since the pronunciation is different.  Another example (of both
\underline{homography} and \underline{homonymy}) is {\it present}
[gift] and {\it present} [the current moment of time].  Note that
the distinction between \underline{homographs} and
\underline{polysemes} is not sharp and sometimes difficult to
make.  There are various boundary situations, e.g. the verb {\it
read} [present] and {\it read} [past] may qualify as homograph,
but the meanings expressed are close to each other.

$\bullet$ \underline{Homonymes}: two words (or characters) that
are simultaneously \underline{homographs} and
\underline{homophones}, e.g. {\it left} [past of {\it leave}] and
{\it left} [opposite of {\it right}].  Some homonymes started out
as \underline{polysemes}, but then developed a substantial
difference in meaning, e.g. {\it close} [near] and {\it close} [to
shut (lips)].

$\bullet$ \underline{Heteronyms}: two homographs that are not
homophones, i.e. they are written in the same way, but are
pronounced differently. Normally, heteronyms have at least two
sufficiently different meanings, indicated by different
pronunciations.

$\bullet$ \underline{Key-word} (key-character): a content word
(character) that characterizes a given text with its specific
subject. The operational definition of a key-word (key-character)
is that in a given text its frequency is much larger than in a
frequency dictionary, which was obtained by mixing together a big
mixture of different texts.

$\bullet$ \underline{Language family}: a set of related languages
that are believed (or proved) to originate from a common ancestor
language.

$\bullet$ \underline{Latent semantic analysis}: the analysis of
word frequencies and word-word correlations (hence semantic
relations) in a text that is based on the idea of hidden (latent)
variables that control the usage of words; see \cite{hofmann} for
reviews.

$\bullet$ \underline{Literal translation}: word-to-word
translation, with (possibly) changing the word ordering, as
necessary for making more understandable the grammar of the
translated text. This notion contrasts to the phrasal translation,
where the meaning of each given phrase is translated. The literal
translation can misconceive idioms and/or shades of meaning, but
these aspects are minor for gross (statistical) features of a
text, e.g. for rank-frequency relation of its words.

$\bullet$ \underline{Logographic} \underline{writing system} is
based on the direct coding of morphemes.

$\bullet$ \underline{Mental lexicon}: the store of words in the
long-time memory. The words from the mental lexicon are employed
on-line for expressing thoughts via phrases and sentences; see
\cite{levelt} for detailed theories of the mental lexicon.
Ref.~\cite{levelt} argues that in addition to mental lexicon
humans contain a mental syllabary that is activated during the
phonologization of a word that was already extracted from the
mental lexicon.


$\bullet$ \underline{Morpheme}: the smallest part of the speech or
writing that has a separate (not necessarily unique) meaning, e.g.
{\it cats} has two morphemes: {\it cat} and {\it -s}. The first
morpheme can stand alone.  The second one expresses the
grammatical meaning of plurality, but it is a \underline{bound
morpheme}, since it can appear only together with other morphemes.

$\bullet$ \underline{Phoneme}: a class of speech sounds that are
perceived as equivalent in a given language. An alternative
definition: the smallest unit that can change the meaning. Hence
normally several different sounds (frequently not distinguished by
native speakers) enter in a single phoneme.

$\bullet$ \underline{Pictogram}: a graphic symbol that represents
an idea or concept through pictorial resemblance to that idea or
concept.

$\bullet$ \underline{Polysemes}: are related meanings of the same
word, e.g.  the English word {\it get} means {\it obtain/have},
but also {\it understand} (= {\it have knowledge}). Another
example is that many English nouns are simultaneously verbs (e.g.
{\it advocate} [person] and {\it advocate} [to defend]).

$\bullet$ \underline{Syllable}: is the minimal phonetic unit
characterized by acoustic integrity of its components (sounds),
e.g. the word {\it body} is composed of two syllables: {\it bo-}
and {\it -dy}, while {\it consider} consists of three syllables:
{\it con-} {\it -si-} {\it -der}. In phonetic languages such as
Russian the factorization of the word into syllables
(syllabification) is straightforward, since the number of
syllables directly relates to the number of vowels. In
non-phonetic languages such as English, the correct
syllabification can be complicated and not readily available to
non-experts. Indo-European languages typically have many
syllables, e.g. the total number of English syllables is more 10
000.  However, 80 \% of speech employs only 500-600 frequent
syllables \cite{levelt}. It was argued, based on psycholinguistic
studies, that the frequent syllables are also stored in the
long-term memory analogously to \underline{mental lexicon}
\cite{levelt}.  The total number of Chinese syllables is much
less, around 500 (about 1200 together with tones)
\cite{levelt,obukhova}. Syllabification in Chinese is generally
straightforward too, also because each character corresponds to a
syllable.

$\bullet$ \underline{Token}: particular instance of a word; a word
as it appears in some text.

$\bullet$ \underline{Type}: the general sort of word; a word as it
appears in a dictionary.

$\bullet$ \underline{Writing system}: process or result of
recording spoken language using a system of visual marks on a
surface. There are two major types of writing systems:
\underline{logographic} (Sumerian cuneiforms, Egyptian
hieroglyphs, Chinese characters) and phonographic. The latter
includes syllabic writing (e.g. Japanese hiragana) and alphabetic
writing (English, Russian, German). The former encodes
\underline{syllables}, while the former encodes
\underline{phonemes}.

\section*{Appendix B: Interference experiments}
\label{inter}

The general scheme of interference experiments in psychology is
described as follows \cite{hoosain,hoosain_1}. There are two
tasks, the main one and the auxiliary one. Each task is defined
via specific instructions. The subjects are asked to carry out the
main task simultaneously trying to ignore the auxiliary task.  The
performance times for carrying out the main task in the presence
of the auxiliary one are then compared with the performance times
of the main task when the auxiliary task is absent. The
interference means that the auxiliary task impedes the main one.

There is a rough qualitative regularity noted in many experiments:
interference decreases upon increasing the complexity of the main
task or upon decreasing the complexity of the auxiliary task.

The most known example of interference experiment is the Stroop
effect, where the main task is to call the color of words. The
auxiliary task is not to pay attention at the meaning of those
words. The experiment is designed such that there is an
incongruency between the semantic meaning of the word and its
color, e.g. the word {\it red} is written in black. As compared to
the situation when the incongruency is absent, i.e. the word {\it
red} is written in red, the reaction time of performing the main
task is sizably larger. This is the essence of the Stroop effect:
the semantic meaning interferes with the color perception.

It appears that the Stroop effect is larger for Chinese characters
than for English words; see \cite{hoosain} for a review. This is
one (but not the only) way to show that getting to the meaning of
a Chinese character is faster than to the meaning of an English
word.

Another known interference phenomenon is the word inferiority
versus word superiority effect. In English these effects amount to
the following \cite{drewno}.

If English-speaking subjects are asked to trace out (and count) a
specific letter in a text, they make less errors, when the text is
meaningless, i.e. it consists of meaningless strings of letters
\cite{chen,review_chin_psi}. This is related to the fact that
English words are recognized and stored as a whole. Hence the
recognition of words|nd moving from one letter to another|
interferes with the task of identifying the letter in a single
word, and the English-speaking subjects make more errors when
tracing out a letter in a meaningful text. This is the
word-inferiority effect.

In contrast, if English subjects is presented a single word for a
short amount of time, and is then asked about letters of this
word, their answers are (statistically) more correct if the word
is meaningful (i.e. it is a real word, not a meaningless sequence
of symbols). This word-superiority effect is understood by noting
that a single word is recalled and/or remembered better due to its
meaning.

In contrast to this, Chinese-speaking adults display the word
superiority effect, when the naive analogy with English would
suggest the word inferiority. They do less errors in tracing out a
given character in a string of meaningful characters, as compared
to tracing it out in a list of meaningless pseudo-characters
\cite{chen}.

A possible interpretation of this effect is that, on one hand, the
definition of Chinese words and their boundaries is somewhat
fuzzy, so that the analogue of the English word-inferiority effect
is not effective. On the other hand, the Chinese sentence is
perceived as a whole, inviting analogies with the English
word-superiority effect.

Note that when the Chinese subjects are asked to trace out a
specific stroke within a character we expectedly (and in full
analogy with the English situation) get that it is easier for
Chinese subjects to trace out the stroke in a meaningless
pseudo-character than in a meaningful character \cite{chen}.

\section*{Appendix C: A list of the studied texts}
\label{texts}

\ \ \ \ 1) Two short modern Chinese texts:

- À¥ÂØéä, {\it K\=un L\'un Sh\=ang} (KLS) by Shu Ming Bi, 1987,
(the total number of characters $N=20226$, the number of different
characters $n=2047$). The text is about the arduous military
training in the troops of Kun Lun mountain.

- °¢QÕý´«, {\it Ah Q Zh\`eng Zhu\`an} (AQZ) by Xun Lu, 1922,
($N=18153$, $n=1553$). The story traces the ``adventures" of a
hypocrit and conformist called Ah Q, who is famous for what he
presents as ``spiritual victories".

2) Two long modern Chinese texts:

- ƽ·²µÄÊÀ½ç, {\it P\'\i ng F\'an de Sh\`\i \ Ji\`e} (PFSJ) by Yao
Lu, 1986, ($N=705130$, $n=3820$). The novel depicts many ordinary
people's stories which include labor and love, setbacks and
pursue, pain and joy, daily life and huge social conflict.

- ˮ䰴«, {\it Shu\v\i \ H\v u Zhu\`an} (SHZ) by Nai An Shi, 14th
century, ($N=704936$, $n=4376$). The story tells how a group of
108 outlaws gathered at Mount Liang formed a sizable army before
they were eventually granted amnesty by the government and sent on
campaigns to resist foreign invaders and suppress rebel forces.

3) Four short classic Chinese texts:

- ´ºÇﷱ¶, {\it Ch\=un Qi\=u F\'an L\`u} (CQF), by Zhong Shu
Dong, 179-104 BC, (Vol.{\bf 1}-Vol.{\bf 8}, $N=30017$, $n=1661$).
A commentary on the Confucian thought and teachings.

- É®±¦´«, {\it S\=eng B\v ao Zhu\`an} (SBZ), by Hong Hui, 1124,
(Vol.{\bf 1}-Vol.{\bf 7}, $N=24634$, $n=1959$). A commentary on
the Taoist thought and teachings. Biographies of great Taoist
masters.

- Îä¾­×ÜÒª, {\it W\v u J\=\i ng  Z\v ong Y\`ao} (WJZ), by Gong
Liang Zeng and Du Ding, 1040-1044, (Vol.{\bf 1}-Vol.{\bf 4},
$N=26330$, $n=1708$). A Chinese military compendium. The text
covers a wide range of subjects, from naval warships to different
types of catapults.

- »¢Áá¾­, {\it H\v u L\'\i ng  J\=\i ng} (HLJ), by Dong Xu, 1004,
(Vol.{\bf 1}-Vol.{\bf 7}, $N=26559$, $n=1837$). Reviews various
military strategies and relates them to factors of geography and
climate.

4) A long classic Chinese text:

- Ê·¼Ç, {\it Sh\v\i \ J\`i} (SJ), by Qian Sima, 109 to 91 BC,
($N=572864$, $n=4932$). Reviews imperial biographies, tables,
treatises, biographies of feudal houses and eminent persons.

\section*{Appendix D: Key-characters of the modern Chinese text
KLS} \label{key_kls}

Here is the list of the key-characters in the Pre-Zipfian and
Zipfian range (Table \ref{tab_7}) of the modern Chinese text,
À¥ÂØéä \ {\it
  K\=un L\'un Sh\=ang} (KLS) written by Shu-Ming BI in 1987. The text
is about the arduous military training in the troops of Kun Lun
mountain.

\begin{table}[ht]
\begin{center}
\tabcolsep0.04in \arrayrulewidth0.5pt
\renewcommand{\arraystretch}{1.25}
\caption{\label{tab_7} Key-characters of the modern Chinese text
À¥ÂØéä \ {\it k\=un l\'un sh\=ang} (KLS).} \vskip 0.2cm
\begin{tabular}{|c|c|c|c|c|c|}
 \hline
No. & Rank &  Character & Pinyin & English & Frequency   \\
\hline
1 & 14 & ºÅ & {\it h\`ao} & horn & 157\\
2 & 32 & ¾ü & {\it j\=un}& army  &  86 \\
3 & 44 & ±ø & {\it b\= \i n} & soldier & 67 \\
4 & 113 & ¶Ó & {\it du\`\i}& troop  &  38 \\
5 & 118 & Áî & {\it l\`\i ng}& command  &  37 \\
6 & 123 & ²¿ & {\it b\`u}& troop  &  36 \\
7 & 152 & Õ½ & {\it zh\`an}& fight/war  &  28 \\
8 & 156 & Ãü & {\it m\`ing}& command  &  28 \\
9 & 180 & ·À & {\it f\'ang}& protect  &  24 \\
10 & 213 & Ѫ & {\it xu\`e}& blood  &  20 \\
11 & 216 & Á¢& {\it l\`\i}& stand straight  &  20 \\
12 & 224 & ¹¦ & {\it g\=ong}& honor  &  19 \\
13 & 225 & ǹ & {\it qi\=ang}& gun  &  19 \\
14 & 252 & ¹Ù & {\it gu\=an}& officer  &  16 \\
15 & 295 & ¹ø & {\it gu\=o}& pan  &  14 \\
16 & 299 & ±£ & {\it b\v ao}& protect  &  14 \\
17 & 300 & ÎÀ & {\it w\`ei}& protect  &  13 \\
18 & 352 & Óª & {\it y\'\i ng}& camp  &  11 \\
19 & 355 & ı & {\it m\'ou}& strategy  &  11 \\
20 & 360 & ÉÕ & {\it sh\=ao}& burn  &  11 \\
21 & 394 & ÁÒ & {\it li\`e}& martyr  &  10 \\
22 & 407 & ÍÅ & {\it tu\'an}& regiment  &  10 \\
\hline
\end{tabular}
\end{center}
\end{table}

\section*{Appendix E: Kolmogorov-Smirnov test}
\label{ap_ks}

The Kolmogorov-Smirnov test (KS test) \cite{Kolmogorov,Nicholls}
is used to determine if a data sample agrees with a reference
probability distribution. The basic idea of the KS test is as
follows.

We need to determine whether a given set $X_1$, $X_2$, ... , $X_n$
is generated by i.i.d sampling a random variable with cumulative
probability distribution $F(x)$ (null hypothesis). To this end we
calculate the the empiric cumulative distribution function (CDF)
$F_n(x)$ for $X_1$, $X_2$, ... , $X_n$:
\begin{equation}
F_n(x)=\frac{1}{n}\sum_{i=1}^{n}I_{X_i\leq x},
\end{equation}
\noindent where $I_{X_i\leq x}$ equals to 1 if $X_i\leq x$ and 0
otherwise. Next we define:
\begin{equation}
D_n= \sup_x|F_n(x)-F(x)|.
\end{equation}
The advantage of using $D_n$ (against other measures of distance
between $F_n(x)$ and $F(x)$) is that if the null hypothesis is
true, the probability distribution of $D_n$ does not depend on
$F(x)$. In that case it was shown that for $n\to\infty$, the
cumulative probability distribution of $\sqrt{n}D_n$ is
\cite{Kolmogorov,Nicholls}:
\begin{equation}
P(\sqrt{n}D_n\leq x)\equiv f(x)=1-2\sum_{k=1}^{\infty}
(-1)^{k-1}e^{-2k^2x^2}.
\end{equation}
For not rejecting the null hypothesis we need that the observed
value of $\sqrt{n}D^*_n$ is sufficiently small. To quantify that
smallness we take a parameter (significance level) $\alpha$
($0<\alpha<1$) and define $\kappa_{\alpha}$ as the unique solution
of
\begin{equation}
f(\kappa_{\alpha})=1-\alpha.
\end{equation}
Now the null hypothesis is not rejected provided that
\begin{equation}
\label{ora} \sqrt{n}D^*_n <\kappa_{\alpha},
\end{equation}
where $\sqrt{n}D^*_n$ is the observed (calculated) value of $D_n$.
Condition (\ref{ora}) ensures that if the null hypothesis is true,
the probability to reject it is bounded from below by $\alpha$.
Hence in practice one takes, e.g. $\alpha=0.05$ or $\alpha=0.01$.

Note however that condition (\ref{ora}) will always hold provided
that $\alpha$ is taken sufficiently small. Hence to quantify the
goodness of the null hypothesis one should calculate the p-value
$p$: the maximal value of $\alpha$, where (\ref{ora}) still holds.
For the hypothesis to be reliable one needs that $p$ is not very
small. As an empiric criterion of reliability people frequently
take $p>0.1$.

\begin{table*}[ht]
\begin{center}
\tabcolsep0.18in \arrayrulewidth0.5pt
\renewcommand{\arraystretch}{1.2}
\caption{\label{tab_8} Kolmogorov-Smirnov test (KS test) for the
fitting quality of our results (texts are defined in Tables
\ref{tab_1} and \ref{tab_2}). In the KS test, $D$ and $p$ denote
the maximum difference (test statistics) and p-value respectively.
$D_1$ and $p_1$ are calculated from the KS test between empiric
data and numerical fitting, $D_2$ and $p_2$ are between empiric
data and theoretical result, $D_3$ and $p_3$ are between numerical
fitting and theoretical result; see section III A. Note that for
making the testing even more vigorous the presented results for
the KS characteristics are obtained in the original coordinates
(\ref{00}); similar results are obtained in logarithmical
coordinates (\ref{kabanyan}) that are employed for the linear
fitting. } \vskip 0.2cm
\begin{tabular}{|c|c|c|c|c|c|c|}
\hline
Texts & $D_1$ &  $p_1$ &$D_2$& $p_2$ & $D_3$ & $p_3$  \\
\hline TF & 0.0418 & 0.865 & 0.0365& 0.939 & 0.0381& 0.912\\
TM & 0.0529 & 0.682 & 0.0562 & 0.593 & 0.0581  & 0.568\\
AR & 0.0564 & 0.624 &  0.0469 & 0.783 & 0.0443& 0.825 \\
DL & 0.0451 & 0.812 &  0.0421 & 0.865  & 0.0472& 0.761 \\
AQZ & 0.0586 & 0.587 &  0.0565 & 0.623  & 0.0601& 0.564 \\
KLS & 0.0592 & 0.578 &  0.0641 & 0.496  & 0.0626& 0.521 \\
CQF & 0.0341 & 0.962 &  0.0415 & 0.863  & 0.0421& 0.857 \\
SBZ & 0.0461 & 0.796 &  0.0558 & 0.635  & 0.0616 & 0.538 \\
WJZ & 0.0427 & 0.852 &  0.0475 & 0.753  & 0.0524& 0.691 \\
HLJ & 0.0375 & 0.923 &  0.0412 & 0.875  & 0.0425& 0.862 \\
\hline
\end{tabular}
\end{center}
\end{table*}

We applied the KS test to our data on the character (word)
frequencies; see section III A. The empiric results on word
frequencies $f_r$ in the Zipfian range $[r_{\rm min}, r_{\rm
max}]$ are fit to the power law, and then also to the theoretical
prediction described in section III C. With null hypothesis that
empiric data follows the numerical fittings and/or theoretical
results, we calculated the maximum differences (test statistics)
$D$ and the corresponding p-values in the KS tests.  From Table
\ref{tab_8} one sees that all the test statistics $D$ are quite
small, while the p-values are {\it much larger} than 0.1. We
conclude that from the viewpoint of the KS test the numerical
fittings and theoretical results can be used to characterize the
empiric data in the Zipfian range reasonably well.



\end{CJK}
\end{document}